# AI-augmented histopathologic review using image analysis to optimize DNA yield and tumor purity from FFPE slides

## Author List & Affiliations

Bolesław L. Osinski,[1] Aïcha BenTaieb,[1] Irvin Ho,[1] Ryan D. Jones,[1] Rohan P. Joshi,[1] Andrew Westley,[1] Michael Carlson,[1] Caleb Willis,[1] Luke Schleicher,[1] Brett M. Mahon,[1] Martin C. Stumpe[1]

1. Tempus Labs, Chicago, IL

Corresponding Author:
Bolesław L. Osinski
600 W Chicago Ave Ste #510, Chicago, IL 60654
(833) 514-4187
bo.osinski@tempus.com

## Competing Interests Statement
All authors were employees and shareholders of Tempus Labs at the time this work was done.

# Abstract

To achieve minimum DNA input and tumor purity requirements for next-generation sequencing (NGS), pathologists visually estimate macrodissection and slide count decisions. Unfortunately, misestimation may cause tissue waste and increased laboratory costs. We developed an AI-augmented smart pathology review system (SmartPath) to empower pathologists with quantitative metrics for accurately determining tissue extraction parameters. SmartPath uses two deep learning architectures, a U-Net based network for cell segmentation and a multi-field-of-view convolutional network for tumor area segmentation, to extract features from digitized H&E-stained FFPE slides. From the segmented tumor area, SmartPath suggests a macrodissection area to enrich tumor percentage. To predict DNA yield per slide, the extracted features are correlated with known DNA yields to fit a regularized linear model (R=0.85). Then, a pathologist-defined target yield divided by the predicted DNA yield/slide gives the number of slides to scrape. Following model development, an internal validation trial was conducted within the Tempus Labs molecular sequencing laboratory. We evaluated our system on 501 clinical colorectal cancer slides, where half received SmartPath-augmented review and half traditional pathologist review. The SmartPath cohort had 25% more DNA yields within a desired target range of 100-2000ng. The number of extraction attempts was statistically unchanged between cohorts. The SmartPath system recommended fewer slides to scrape for large tissue sections, saving tissue in these cases. Conversely, SmartPath recommended more slides to scrape for samples with scant tissue sections, helping prevent costly re-extraction due to insufficient extraction yield. A statistical analysis was performed to measure the impact of covariates on the results, offering insights on how to improve future applications of SmartPath. Overall, the study demonstrated that AI-augmented histopathologic review using SmartPath could decrease tissue waste, sequencing time, and laboratory costs by optimizing DNA yields and tumor purity.

# Introduction

Next-generation sequencing (NGS) has become an integral technique in the molecular diagnosis, prognosis, and treatment of cancer. To properly assess tumor tissue with NGS, solid samples must be dissected to meet minimum DNA input and tumor purity requirements [1,2,3,4,5]. In standard practice, pathologists visually inspect hematoxylin and eosin (H&E)-stained, formalin-fixed, paraffin-embedded (FFPE) slides to determine how much tissue should be dissected and whether macrodissection is necessary to enrich for tumor cells [6,7]. Besides meeting minimum input requirements, pathologists must also avoid recommending excessive dissection as tumor tissue is valuable and may be needed for further molecular tests. Tissue stewardship guidelines can help pathologists achieve this balance between sufficient and excessive dissection [4]. However, following these suggestions using manual dissection techniques is difficult, and thus, there is an increasing need to optimize tissue extraction procedures as NGS becomes more relevant in clinical practice.

NGS pipelines have undergone tremendous advancements in the past decade [8,9,10], including the development of automated dissection systems for tissue extraction. Laser-capture microdissection was introduced about two decades ago [11,12,13], but has not been widely adopted in clinical laboratories because precise dissection of single tumor cells from FFPE slides is rarely necessary for clinical testing [14]. Lower resolution mechanical microdissection systems have also been developed as more clinically pragmatic alternatives [15,16,17,18]. These systems can be combined with digital slide marking (digitally guided macrodissection), enabling integration with computer vision models for tumor enrichment [19,20]. However, even automated microdissection systems rely on visual estimation by a pathologist to determine how many slides should be scraped. Unfortunately, consequences of visual misestimation include sequencing failure, tissue waste, and increased laboratory costs and turnaround times.

Advancements in artificial intelligence (AI)-augmented pathology have largely focused on computer-aided diagnostic systems, which can increase accuracy, save time, and reduce labor [21,22]. One study found that AI-assisted pathologists had significantly higher sensitivity and reduced review times when detecting breast cancer metastasis in lymph nodes [23]. In another study, an AI-augmented microscope that overlays tumor classification results over the microscope view in real time assisted pathologists in detecting metastatic breast cancer and prostate cancer with high accuracy [24]. Although the field of AI-assisted pathology review is growing quickly, no recommendation systems exist for tissue stewardship in molecular pathology NGS testing.

Here, we developed SmartPath: a computer vision-based method to empower pathologists with quantitative metrics, allowing them to accurately determine tissue input parameters for desired DNA yields. Echoing design principles of AI-assisted pathology outlined by others [25], SmartPath functions as a pathologist-in-the loop system rather than a standalone predictor. These predictions are displayed in a browser-based user interface (UI) viewed during pathology review. We tested SmartPath in an internal trial to assess the impact of AI-augmented pathology review on DNA extraction and NGS workflow costs in colorectal tumor samples from a real-world clinical setting. We quantify immediate impacts of the AI-assistance on tissue usage and the extracted DNA content, as well as on two NGS workflow costs: the total number of extractions attempted and the DNA extraction-to-sequencing time (T-seq). A thorough statistical investigation of the impact of clinical covariates on these metrics was conducted, revealing factors that influence NGS success beyond tissue input parameters alone.

## Methods

# Model development

Before evaluating the SmartPath AI-augmented pathology review system in an internal trial (see Methods: Internal Model Evaluation Trial), we first developed the model through extensive validation experiments. The models used can be grouped into two categories: feature generation and DNA yield estimation. Feature generation aims to extract features from a single H&E-stained histopathology whole-slide image (WSI). These features are then used to train a DNA yield estimation model that automates pathologist tissue quantity selections to achieve a total extracted DNA mass within a target range. This modeling pipeline is summarized in Figure 1A.

## Feature Generation Pipeline

Feature extraction relies on pretrained tumor and cell segmentation models (Section S1). The tumor segmentation, based on a multi field-of-view network with a fully convolutional ResNet-18 backbone [26], produces segmentation maps of tumor- and lymphocyte-rich areas (Section S1.1). The cell segmentation model, based on the U-Net architecture [27], produces segmentations of individual tumor nuclei and lymphocytes throughout the whole image (Section S1.2). The tumor and cell segmentation models are combined to assign identities to tumor and lymphocyte cells (Section S1.3). Model outputs are fed to our feature generation pipeline, which extracts four feature groups: cell counts, tumor shape, cell nucleus shape, and cell nucleus texture, totaling 3,461 features from each slide (Section S2).

*Macrodissection area masking during training*

Tissues containing a large proportion of non-tumor tissue required macrodissection to enrich the percentage of tumor cells extractable by scraping. Before DNA extraction, the macrodissection area was estimated by a pathologist to include an area containing ≥20% tumor cells, and was hand drawn onto the slide. This area was later manually scraped by a technician

Approximately 30% of the training samples were macrodissected. To ensure the computed imaging features from these slides were only from the hand-drawn region, we developed an ink detection model which was post-processed to produce a macrodissection area mask (Section S3).

*Macrodissection area masking during inference*

During inference, the pathologist had not yet drawn the macrodissection area, and instead it was estimated algorithmically. We predicted an optimal macrodissection area using the predictions of our tumor segmentation model. The predicted tumor area was converted to a binary mask and post-processed as follows: small, isolated regions less than 1/10 of the largest tissue size were removed and holes within each remaining tumor area were filled. The resulting mask was then downsampled to 0.625x magnification and dilated (3x3 kernel, 25 iterations) to produce an area fully enclosing the tumor. The dilation parameters were manually chosen to produce a contour mimicking hand-drawn macrodissection areas. However, this procedure sometimes produced masks with too many separate contours as compared to the hand-drawn areas. When there were >3 separate contours after dilation, we applied a convex hull to encircle them. Because this method is essentially a dilation of the main tumor area, the resulting area usually contained >20% detected tumor cells. Details on the validation of this method are in Section S4.

## Training and validation sets for DNA yield prediction

The core model underlying SmartPath is the prediction of DNA yield per slide using linear regression on extracted imaging features. To acquire a training set, the Tempus database was searched for slides scanned between January 2018 and January 2020 containing lung, breast, or colorectal cancer (CRC) primary tumor tissue. Three different cancer types were used for training because the average nuclear content per cell should be similar across cancers, and therefore extracted cell-count features should follow the same linear correlation with DNA yield regardless of cancer type. We confirmed this assumption by estimating the yield per cell for each cancer type (measured DNA yield per slide / predicted number of cells per slide) and finding no statistical difference between the means. Using the three cancer types also increased the training set size, which might help the model better generalize to unseen data in the future.

Aspirates and cytology specimens were excluded from the training set, as were slides with no recorded DNA mass or scraping, leaving a final training set of 1,605 slides. Approximately 28% were previously macrodissected, reflecting the rate at which samples are macrodissected in our normal clinical workflow. Each slide was inspected to ensure they had hand-drawn macrodissection areas. Characteristics of the training set are shown in Supplementary Table 1.

We also acquired a separate validation set of 332 retrospective samples from the same database, restricted to only CRC tissues, which was used for selecting a model with the best performing parameter combination (Supplementary Table 2). The validation set was enriched for macrodissected cases (57% were macrodissected) to ensure thorough evaluation of our macrodissection estimation algorithm. Characteristics of the validation set are shown in Supplementary Table 1.

## DNA Yield Prediction

### Ground truth definition for DNA yield prediction model

The ground truth for training the DNA yield prediction model was taken as the extracted DNA yield from FFPE slides (see Section S5 for details on DNA extraction procedure). Each slide in the training and validation sets was an archival H&E slide representative of the unstained slides already extracted and sequenced by our NGS laboratory. Although most underwent only one

DNA extraction attempt, some had multiple, in which case the imaged slide may have been closer to the tissues used for the 2nd extraction attempt. Therefore, the ground truth was defined as:

(DNA1 yield) / ($N_{slides}$), *if only 1 extraction attempted*
(DNA1 yield + DNA2 yield) / ($N_{slides}$ for DNA1 + $N_{slides}$ for DNA2), *if >1 extractions attempted*

Section S6 contains more details on the ground truth definition.

*Parameter and feature exploration for final model selection*

For parameter exploration we used the full feature set (3,641 features). Because the number of features was larger than the number of samples in the training set (1,605), the linear model severely overfit and failed to generalize without regularization (Figure 2A,B). To ameliorate this issue, optimal regularization parameters were determined by parameter sweeping across L1 and L2 regularization strengths. Each regularization was tested with natural log and Box-Cox power transformations on features and ground truth. The parameter combination with the best validation set performance ($R$=0.818) was a log transform and an L1 regularization with strength=0.01 (see Supplementary Table 2 for full parameter exploration). Predictions of this optimal model are plotted against training and validation ground truths in Figure 2C,D.

To confirm that including all 3,641 features was advantageous, we performed a 200-fold cross validation using an 80/20 train/val split of the training set using the optimal parameter combination found from the exploration. We measured the mean coefficient magnitude across folds for each parameter. The top 10 features accounted for 93.4% of the model coefficient magnitude and were from a combination of feature groups (cell counts, tumor shape, cell shape, texture), with the total cell count having the highest importance (Supplementary Figure 2). Further confirmation of the usefulness of keeping all features was done in a 200-fold cross validation experiment where cumulatively more features were included. Starting with just 1 feature, then progressively adding features by group, performance on the cross-validated training set and the withheld validation set increased as more features were included (Supplementary Figure 3). This confirmed that inclusion of all features gave the best performing model.

We also explored the inclusion of categorical features, such as procedure type, tissue site, and institution. However, inclusion of categorical features did not offer any significant boost in performance of cross-validated models, and their model coefficients were consistently pushed towards 0 by regularization. Because the final model was to be run in a real-world scenario, where image artifacts may cause some features to have infinite or non-numeric values, additional steps were also taken to ensure that such situations were handled smoothly in the inference pipeline (Section S7).

## Target yields for number of slides prediction

The goal of SmartPath is to recommend the number of scraped slides needed to achieve a DNA yield between 100-2000 ng. To convert the predicted DNA yield per slide into a recommendation of how many slides to scrape, we divided a target yield by the predicted yield per slide and rounded down to the nearest integer. This target yield is a tunable operating point of the algorithm. During the trial, the SmartPath system presented the number of slides needed to achieve a target yield of at least 100ng, 400ng, or 1000ng. For details on how these target yields were selected, see Section S8.

We chose three target yields instead of one to give pathologists more flexibility. Because the relationship between number of slides scraped and DNA yield is linear, pathologists can also use the target yields to interpolate the recommended number of slides if they choose. This design choice emphasizes the principle of AI-augmented decision making, rather than AI-automation.

# Internal model evaluation trial

## Trial design

To test the viability of our system in practice, we undertook an internal trial using clinical CRC samples to evaluate SmartPath compared to traditional (Trad) pathologist review. Trad and SmartPath workflows are summarized in Figure 1B. Sample sizes of 250 SmartPath and 250 Trad samples were determined by power analysis at significance level=0.01 and power=0.8 (Section S9). The internal trial was designed to be run in tandem with standard clinical workflow, mirroring every step until pathologist review. Before pathologist review, each FFPE block was cut into 20 sections and affixed to glass slides. One slide midway through the levels was stained for H&E and designated for pathologist review, while the others were designated for scraping. If the tumor was CRC primary and met additional inclusion criteria (Supplementary Table 3), the slides were flagged for trial enrollment and were aggregated separately to avoid mixing with the rest of the clinical workflow. Once the enrolled samples for the day were verified, they were assigned in alternating order to Trad or SmartPath cohorts and assignments were recorded in a log, aiming to collect roughly equal numbers per day.

The Trad cohort H&E slides were a control group of samples which passed through our established pathology review workflow. All samples were still reviewed by a pathologist in a timely manner to not disturb the existing clinical sequencing workflow. Samples assigned to the Trad cohort were re-entered into the clinical workflow and path reviewed that same day.

The SmartPath cohort H&E slides were scanned on the Philips Ultra Fast Scanner (Philips, Eindhoven, The Netherlands) to produce a digitized WSI at 40× base magnification level (0.25 µm/pixel). Slide scanning automatically triggered classification and DNA-yield prediction on the WSI (Section S10). To minimize interference with existing clinical workflow, SmartPath-assisted review was conducted the morning after scanning, although in principle same day review is quite feasible because scanning and model deployment take only minutes to complete. After pathology review, the recommended number of slides were scraped, DNA was extracted (Section S5), and next-generation sequencing was conducted using the Tempus xT platform [8]. This process was repeated daily over the span of several months until 251 SmartPath and 250 Trad cohort samples were accumulated.

During SmartPath-assisted review, the pathologist viewed a custom-built UI (Supplementary Figure 5) displaying recommendations for macrodissection area and number of slides needed to achieve at least three possible target DNA yields of (100 ng, 400 ng, or 1000 ng). The pathologist had the option to accept or reject model recommendations. If they chose to accept, the desired target yield was selected in the UI. If they disagreed, they recorded the rejection in the trial log and indicated if the rejection was driven by clinical reasoning or model performance. In cases where the pathologist disagreed with the predicted macrodissection area, the pathologist drew their own microdissection area and the case was marked as a deviation.

## Internal trial performance evaluation metrics

The impact of AI-assistance in the trial was evaluated by several metrics from two main categories: extraction metrics and NGS workflow costs.

Extraction metrics
- Percent DNA yield within range: We defined the DNA yield target range as 100-2000ng. The minimum of 100ng was chosen to reflect the minimum input quantity of DNA requested by most NGS laboratories, which is between 50-200ng [2,3]. The maximum of 2000ng was chosen by collaborating pathologists as a reasonable cutoff indicating surplus of extracted DNA. The percent of samples below this range (<100ng) is referred to as undershoot, and above this range (>2000ng) is referred to as overshoot.
- Number of slides scraped (N slides): This measures the number of slides scraped for 1st DNA extraction attempt. We did not count slides scraped for later extraction attempts because additional attempts did not receive AI-assistance. This metric also excludes any slides scraped for RNA extractions or scraped after first successful DNA-seq.

NGS workflow costs
- Extraction count: This metric counts the number of extraction attempts made towards the first DNA-seq attempt. It excludes extractions for RNA and extractions made after the first DNA-seq attempt.
- T-seq: DNA extraction-to-sequencing time, defined as the time elapsed from the first extraction attempt to the first successful DNA-seq attempt (Supplementary Figure 6). This includes only the time period that can be influenced by AI-assistance, excluding RNA-sequencing and any subsequent DNA sequencing. This definition is restricted to the context of this study and is not reflective of Tempus' operational turnaround time.

For each of these metrics, we also present the effect of two effect modifiers, the tissue area and extraction quality.

Effect Modifiers

- Tissue area: Pathologists partially rely on tissue area to estimate how many slides should be scraped for extraction, where small tissues tend to have more slides scraped than large tissues. We split the tissue area at the 50th percentile of the distribution (85.46 mm$^2$) for samples enrolled into the trial to obtain two groups: large (mean area 285.29 mm$^2$) and small (mean area 18.53 mm$^2$) (Supplementary Figure 7A).
- Extraction quality: An in-house measure similar to other established methods [28] for evaluating the quality of extracted DNA. Briefly, the extracted nucleic acid is assessed with a Fragment Analyzer (Advanced Analytical Technologies, Ames, IA), which produces a distribution of nucleic acid fragment lengths, measured in base pairs. The fragment length distribution is split into custom-defined ranges corresponding to short, intermediate, and long fragments and the amount of the distribution in each of these ranges is quantified. To produce the extraction quality, the fragment data is combined with extracted DNA mass and binned into three quality levels: low, intermediate, or high. Low-quality samples generally have short fragments and low DNA yield (usually <90ng), while high-quality samples generally have long fragments and high DNA yield (usually >400 ng).

Although small tissue samples can have high-quality DNA fragmentation, for most samples small tissue area correlates with low extraction quality (Supplementary Figure 7B).

### Internal trial data quality control

A total of 501 samples were enrolled into the trial, with 18 rejected at pathology review, four erroneously enrolled either with incorrect cancer type or procedure type, and one with an incorrect indication of number of slides scraped. Two more samples were also removed because their sequencing was delayed due to human error. This left 476 samples for the overall analysis (233 Trad, 243 Smart). For analysis of T-seq, an additional 16 samples were dropped because they did not reach DNA-seq, and therefore did not have a defined sequencing time interval, leaving 458 samples (226 Trad, 232 Smart).

## Methods: Statistical analysis of covariates

The FDA guidance for adjustment for clinical covariates in clinical trials (Docket number FDA-2019-D-0934) advises experimenters to identify the covariates expected to have an important influence on the primary outcome. The primary outcomes for the present work are the DNA extraction yield and workflow costs. Extraction yield may depend on sample age, as older samples may suffer from nucleotide degradation [29]. It may also depend on the individuals involved in the extraction (i.e., the pathologist and the technicians). The day on which the sample is extracted could have an impact on workflow turnaround time due to weekly lab scheduling cycles. We included these sample-level measures as covariates (Table 1) and also recorded several patient-level characteristics which are commonly reported in cancer studies (Supplementary Table 4).

## Chi-squared test for covariate imbalance

Covariate imbalance between treatment groups can be a source of bias skewing the effect of the treatment on outcome variables. To measure imbalance between SmartPath and Trad cohorts for the sample and patient-level characteristics (Table 1, Supplementary Table 4), we performed a chi-squared test on the contingency tables of each covariate. Contingency tables were computed by cross-tabulating counts for each characteristic and chi-squared tests were performed in Python 3.7 using scipy.chi2_contingency [30].

## Analysis of Covariance using Generalized Linear Models

An analysis of covariance (ANCOVA) allows researchers to dissociate contributions of additional covariates from the treatment to the total variance. For the present application the treatment variable is the trial cohort (Trad or SmartPath) and the dependent variables are the following outcome metrics: DNA mass undershoot boolean (1 if <100ng, 0 otherwise), number of slides scraped (N slides), extraction count, and T-seq. Traditional ANCOVA is designed to run on normally distributed samples assuming linearity and homoscedasticity (constant variance across residuals). However, most of these metrics are not normally distributed, and thus appropriate distributions were chosen to model these dependent variables with generalized linear models (GLMs) [31]. All ANOVA analyses were performed in R version 0.4.4 [32]. For details on GLM selection see Section S11.

### Scaling and encoding of covariates for GLMs

Covariates had to be appropriately encoded for the analysis. Trial cohort was dummy encoded as a binary indicator (0 - SmartPath, 1 - Trad). Extraction quality was numerically encoded as ordinal variables (0, 1, 2). Extraction day-of-week was encoded numerically from Monday to Sunday as 0-6. Sample age was log-transformed. Procedure type, pathologist, and extraction tech were dummy encoded, dropping one category from each to eliminate correlations. For more justification of these encoding choices see Section S12.

### Univariate and multivariate GLMs

GLMs were fit only for the subset of the samples most in need of AI-assistance, namely small tissues with low extraction quality. Univariate GLMs were initially fit using the sample-level (Table 1) and patient-level characteristics (Supplementary Table 4) as independent variables, but significant effects were not found for any of the patient-level characteristics. Multivariate models for each metric are built using only those variables with significant association in univariate tables (Supplementary Tables 5–8). For details on construction of univariate and multivariate models, see Section S13.

# Results

## Impact on extraction metrics

### AI-assistance improved DNA yield within a target range of 100-2000ng

The fraction of samples within the target range was significantly improved for the SmartPath cohort (Trad=0.56 +/- 0.064 vs SmartPath=0.70 +/- 0.058, $P$=0.005, a 25% increase, Figure 3A). This was primarily due to limiting over-extraction, as the fraction of samples with mass that overshot the desired range was also significantly improved (Trad=0.32 +/- 0.06 vs SmartPath=0.18 +/- 0.049, $P$=0.001, a 14% decrease, Figure 3A). The fraction of samples that undershot the desired range was not improved overall.

Tissue characteristics, such as tissue area and DNA fragment quality, are known to impact tissue extraction [33,19]. We confirm that these effects exist in our data as well. In Figure 3B, we subset the data into large and small tissue area groups (defined in Methods), revealing that reduction in overshoot was restricted to large tissues. When further subset by extraction quality (defined in Methods), the reduced overshoot effect was primarily seen in large tissues with high extraction quality (Figure 3C, top). Therefore, AI-assistance helped pathologists preserve tissue use for samples that were already likely to succeed NGS. Subsetting also revealed a trend in reduction of the undershoot fraction for small tissues with low extraction quality (Figure 3C, bottom). Although the difference was not significant (chi-squared $P$=0.088), there were only 50 samples in this subset and the sample size may be underpowered to measure the effect. As discussed in subsequent sections, however, subsetting by small tissue area and low extraction quality showed significant improvements in other metrics.

Laboratory costs associated with overshoot are generally lower than costs associated with undershoot; an overshoot is a waste of tissue, but an undershoot may require re-extraction which is a waste of materials, time, and tissue. In other words, there is a trade-off between tissue stewardship and prevention of re-extraction. Our model design intention was to bias away from tissue stewardship in favor of preventing re-extraction, but evidently it was not biased enough. In retrospect, we note that target yields of the model could have been tuned even higher to improve the undershoot fraction at the expense of the overshoot fraction. This possibility was explored in a simulation (Supplementary Figure 9). According to the simulation, had we scraped 1.4x more slides (in essence, scaling the target yields from 100ng, 400ng, and 1000ng to 140ng, 560ng, and 1400ng) the undershoot fraction for the SmartPath cohort could have dropped below that of the Trad cohort, while still maintaining a reduction in overshoot fraction.

## AI-assistance fosters more efficient use of tissue slides

Across all CRC samples from the trial (n=476), DNA yields <100ng almost always resulted in multiple extraction attempts (Supplementary Figure 10). These results demonstrate the importance of better metrics for scraping parameters, as more slides should be scraped initially when lower DNA yields are expected to avoid repeating extraction. While NGS laboratories typically scrape 5-10 FFPE slides per extraction [34], our AI model recommended a broader distribution of slides for scraping compared with the Trad cohort (Figure 4A). Large tissues in the SmartPath cohort usually had only one or two slides scraped, thus conserving tissue in this subset. On the other hand, small tissues in the SmartPath cohort usually had >10 slides scraped (Figure 4A,B). Therefore, while the mean number of slides was not significantly different between SmartPath and Trad cohorts across all tissue sizes (Figure 4B), slides in the SmartPath cohort were used more efficiently. However, because the distribution of N slides was not normal, the median should also be considered. While the overall mean number of slides scraped per sample in the SmartPath cohort was slightly higher than in the Trad cohort (7.7 $\mp$ 5.91 SmartPath vs 7.62 $\mp$ 3.0 Trad), the median in the SmartPath cohort was lower than in the Trad cohort (6 SmartPath vs 10 Trad).

      Further subsetting the data by extraction quality also shows that SmartPath recommended fewer slides for large tissues regardless of extraction quality (Figure 4C, low quality $P$=0.07, intermediate quality $P<<0.01$, and high quality $P<<0.01$). Although there was no significant difference, an opposite trend was observed for small tissues with low and intermediate extraction quality (Figure 4C bottom), where more slides were recommended in the SmartPath cohort. For the subset of small samples with high extraction quality, the SmartPath and Trad cohort means were very similar. This could be desirable as high-quality samples are already likely to succeed NGS.

# Impact of AI-assistance on NGS workflow costs

## Number of extraction attempts is similar between cohorts

The number of extraction attempts is an important metric for workflow improvement because re-extractions are financially and temporally costly. The distributions of N extractions for SmartPath and Trad cohorts are shown in Figure 5A. As these distributions were not normal, they are represented with Poisson distributions for calculation of statistics (see Methods). The distributions are dominated with cases with only one extraction, which is already the optimum. Overall, no significant difference in mean number of extractions per sample was observed between the SmartPath and Trad cohorts Figure 5B. Grouping by extraction quality reveals that all high-quality samples were already performing at optimum for this metric, with only one extraction per sample, and intermediate-quality samples were performing near optimum. On the other hand, the SmartPath cohort had a decreased mean extraction count for low-quality samples, albeit not significant (Poisson rate ratio $P$=0.212).

      Further subsetting by tissue area reveals that the decrease in mean extraction count per sample approaches significance (Poisson rate ratio $P$=0.052) for low-quality samples with small

tissue area (Figure 5C, bottom). This result suggests that AI-assistance may be useful in preventing re-extractions for low quality samples with small tissue area, which are the samples most in need of improvement. However, there was a significant increase in mean extraction count for large intermediate-quality samples in the SmartPath cohort (Poisson rate ratio $P$=0.017), caused by four samples which had >1 extraction count. The higher extraction count may be partly due to the age of these samples. SmartPath cohort samples were on average older than the Trad cohort (Supplementary Figure 12A), and for large samples in the SmartPath cohort, those with intermediate quality were also the oldest (Supplementary Figure 12B). Older samples correlate with higher extraction count (Supplementary Figure 12D), likely because they tend to be more degraded.

### AI-assistance reduced DNA sequencing time for low quality samples with small tissue areas

Figure 6A shows the distribution of T-seq in the SmartPath and Trad cohorts. Similar to extraction counts, there was no significant difference in the mean T-seq between the two cohorts (Figure 6B; Trad 3.74 +/- 1.67 days, SmartPath 3.89 +/- 1.67 days). We expect T-seq to follow a similar trend as extraction count because they are strongly correlated (Supplementary Figure 11). High-quality samples showed almost no difference in T-seq between cohorts, a reflection of the fact that extraction count is already optimal for high-quality samples. Intermediate quality samples showed a significant increase ($P$=0.018) for the SmartPath cohort, likely due to the same samples that drove up extraction count for this group. However, when subset by tissue area the T-seq for small low-quality samples was almost 2 days shorter in the SmartPath cohort compared with Trad (Figure 6C, bottom; Trad 6.90 +/- 2.77 days, SmartPath 4.97 +/- 2.06 days, $P$=0.025).

## Univariate analysis of covariates on full trial dataset

To determine if the effects observed were due entirely to the experimental condition (SmartPath vs. Trad) alone, we considered covariates of the study and identified several variables, as detailed in Table 1. Despite an attempt to randomize samples by alternating assignment to SmartPath and Trad cohorts each day (see Methods), imbalances were detected. Moderate imbalance was detected for sample age (chi-squared $P$=0.06), while strong imbalance was detected for pathologist (chi-squared $P$<1e-60), extraction day-of-week (chi-square $P$=0.0008), and extraction tech (chi-square $P$=0.0001).

While dataset imbalance in covariates suggests other sources of variability besides the experimental condition, it alone does not prove that they impacted the dependent metrics. With univariate GLMs (see Methods), we quantified the correlation each covariate had with the following four trial metrics: DNA Mass undershoot boolean (True if <100 ng), number of slides scraped, extraction counts, and T-seq. Summary statistics of univariate GLMs for each covariate are presented in Supplementary Tables 5-8.

For three of the trial metrics, undershoot Boolean, number of slides scraped, and extraction count (Supplementary Tables 5-7), the extraction quality ($P < $ 2e-16, $P=$5.96e-05, $P < $ 2e-16, respectively) and tissue area ($P=$4.01e-06, $P < $ 2e-16, $P=$2.78e-06, respectively) were more predictive than any of the covariates. These same three metrics also were significantly correlated with procedure type. This is expected, as procedure type is strongly correlated with tissue area, where needle biopsies tend to have much smaller area than resections.

However, for T-seq (Supplementary Table 8) the extraction day-of-week was the most predictive variable ($P<$2e-16). This is a known effect, where sequencing times for samples extracted later in the week tend to be longer than samples extracted earlier in the week (Supplementary Figure 13B, see Discussion). Despite the randomized trial design, samples in the SmartPath cohort tended to be extracted later in the week than samples in the Trad cohort (Supplementary Figure 13A).

For all metrics except for N slides, the extraction tech group univariate GLMs showed significance ($P<$0.05). Although this variable has high cardinality (25 categories), the Akaike Information Criterion (AIC) in these cases was lower than some of the other variables (Supplementary Tables 5-7), suggesting that these were not random correlations. Imbalance in the extraction tech group was also meant to be eliminated by trial design but persisted despite our efforts.

Only one of the metrics, N slides, was significantly predicted by sample age. None of the metrics had strong correlation with pathologist. In the next section, we investigate the impact these covariates had on the main effect of the experimental variable, the trial cohort.

## Multivariate analysis of covariates for subset of samples with small tissue area and low extraction quality

In Figures 3 - 6 we identified that AI-assistance was most effective for samples with small tissue areas and low extraction quality. As this is the most interesting subset of samples, we restricted the following multivariate analysis to this subset (N=50). For modeling each outcome metric, we chose only those covariates significantly associated ($P<$0.05) with the outcome metrics in the univariate analysis (Supplementary Tables 5-8). We excluded procedure type, as it is already strongly correlated with tissue area and no surgical resections are present in this subset. We also excluded the extraction tech group as this variable has very high cardinality (25 categories) relative to the number of samples in this subset.

To measure the influence of these covariates on the main effect of AI-assistance, we compared a univariate GLM using trial cohort as the independent variable, to the multivariate GLMs for each outcome metric (see Methods). In the interest of brevity and focus, we limited the multivariate analysis to only 4 outcome metrics. In particular, we did not include the target and overshoot fractions as outcome metrics in the present covariate analysis, but instead only considered the undershoot (as defined in Figure 3). Summary statistics of these GLMs are shown in Table 2.

For the small tissue area and low extraction quality data subset, the trial cohort alone was significantly predictive for both extraction count and T-seq ($P=$0.049 and 0.026, respectively). The other two metrics, undershoot and N slides, had univariate associations with trial cohort approaching significance ($P$=0.055 and 0.075, respectively). In all outcome metrics, the inclusion of covariates in multivariate GLMs raised the trial cohort p-values, suggesting that the main effect can be partially explained by the covariates. However, the increase in trial cohort p-value was moderate.

Inclusion of covariates increased the AIC for multivariate GLMs of undershoot from 54.57 to 55.04 and N slides from 65.52 to 74.11, indicating that covariates carry little additional information about these metrics, adding complexity without proportionally improving the fit. For both NGS workflow metrics, though, the AIC was reduced (extraction count: from 120.21 to 118.91, T-seq: from 51.90 to 45.73), indicating that the covariates improved the model without adding unnecessary complexity.

The only significantly predictive covariate in the multivariate GLMs was extraction day-of-week, which strongly associated with T-seq ($P$=0.006). The correlation of extraction day-of-week with T-seq is a known effect in our NGS laboratory (Supplementary Figure 13). Although the cohort imbalance in day-of-week was meant to be eliminated by alternating assignment of samples to SmartPath and Trad cohorts each day (see Methods), unfortunately the imbalance persisted (Table 1). Overall, the multivariate analysis shows that the covariates considered here have a measurable impact on the main effect of trial cohort. The adjusted effect of trial cohort is weaker upon inclusion of covariates, however it is difficult to ascertain if this fully explains the main effect as the main effect is itself underpowered (N=50). Future trials of this tool should ensure that sufficiently high numbers of samples are available in the small tissue and low extraction quality regimes and should more rigorously control for covariate imbalances.

# Discussion

Here, we developed SmartPath, a computer vision tool to assist pathologists in determining NGS tissue input parameters and tested this tool in a real-world clinical setting. Compared to the group that received traditional pathology review, AI-assistance produced significantly more DNA yields falling within a target range of 100-2000 ng. The AI-assisted model also improved tissue stewardship by recommending scraping of more slides for samples with small tissue areas, but fewer slides for samples with large tissue area.

Although we hoped to see improvements in two NGS workflow costs, the extraction count and T-seq, no significant difference was found between the full populations of SmartPath and Trad cohorts for either cost. Notably, though, the similar extraction count and T-seq indicates that improvements in tissue stewardship were not made at the expense of these costs. Furthermore, it is known from the literature that tissue size and quality influence NGS success. In particular, NGS fails more often for smaller samples and those with poor fragmentation quality [33,19]. When subsetting the data by sample size and quality, we observed a significant reduction in T-seq for small, low-quality samples. Additionally, we found that the subset of high-quality samples had only one extraction attempt regardless of SmartPath or Trad treatment, which is already optimal, and therefore the extraction count cost could not be improved for these samples. This subset also represented a majority of the samples in the entire trial, while only ~13% were low quality. Therefore, lack of overall improvement in NGS workflow costs is largely due to over-representation of high fragmentation quality samples in our cohort.

Colorectal cancer was chosen for the trial for internal workflow considerations, but the present algorithm can be trivially generalized to other cancer types by replacing the underlying tissue and cell segmentation models with tissue-specific models. Other cancer types, such as non-small cell lung cancer and especially pancreatic cancer, have higher rates of low fragmentation quality samples and may benefit even more from the AI-augmented pathology review system (Supplementary Figure 15).

Proper randomization is necessary to eliminate all biases in trials, but for many real-world trials like ours, this is not possible due to external constraints. One bias identified was the strong influence of extraction day-of-week on T-seq, likely due to weekly batch effects and staffing cycles which cause sequencing times to be longer for samples extracted later in the week. This resembles a well-documented weekly phenomenon in healthcare, termed the "weekend effect" [35]. The trial also took place in Aug-Dec 2020 durring the lockdown period of the COVID-19 epidemic, and thus effects due to limited personnel were likely exaggerated. This bias was meant to be eliminated by the trial design by enrolling similar numbers of samples into SmartPath and Trad cohorts per day, but multivariate analysis showed that the AI-assistance effect was reduced after inclusion of extraction day-of-week as a predictor (Table 2).

Sample age has a known effect on FFPE sample extraction success. Evidence exists suggesting that samples >7 years old are unsuitable for NGS [29]; however, samples much older than that have been successfully sequenced [36]. Exclusion of such older samples can negatively impact patients' lives, so our laboratory does not reject samples for sequencing due to old age. By chance, the SmartPath cohort samples were generally older than Trad samples (Supplementary Figure 12), and therefore a bias towards sequencing failure could have been introduced for the SmartPath cohort.

Imbalance was also detected when treating pathologist as a covariate. The SmartPath cohort was mostly reviewed by one pathologist who was trained to use the UI at the start of the trial. This was an operational constraint due to limited resources that were split between normal lab operation and conducting this trial. Our top priority was not to disturb our existing NGS clinical workflow. Despite this imbalance, the pathologist identity did not strongly associate with any of the outcome metrics (Table 2). In fact, the mean values of the outcome metrics were similar between the pathologist that reviewed the largest number of SmartPath cohort samples and the pathologist that reviewed the largest number of Trad cohort samples (Supplementary Figure 14).

Overall, our statistical analysis of covariates highlights various sources of bias that affect the deployment of NGS workflow improvement tools in real-world settings. Future tissue recommendation models could improve upon the current work by taking these sources of bia into account as variables in the model itself and/or explicitly eliminating these effects through trial design.

Along with eliminating biases, several improvements to the existing DNA yield prediction strategy could also be made. The operating point determination was based on varying the threshold for target yield, but the slope of the linear model could also be altered to bias towards scraping more slides for lower yields while scraping fewer sides for higher yields. Although feature extraction relied on neural networks, DNA yield prediction was accomplished with a relatively simple modeling approach using a regularized linear model on extracted features (primarily cell counts). This was done because the extracted features are easily interpretable to pathologists and because the relationship between number of slides and extracted DNA yield is inherently linear. However, future approaches may improve results by predicting DNA yield directly with a neural network.

While the present model relies mostly on extrinsic features such as cell counts and tumor area for correlation with tumor yields, more work could be done to investigate intrinsic features, such as slide preparation quality, and possibly even fragmentation quality with imaging means. An imaging-based predictor could potentially be trained to provide a prior on sample quality. Training data for an imaging-based quality predictor could come from a combination of established DNA quality metrics, including fragment analyzer data, qPCR assays to measure the amount of amplifiable DNA in a sample, the DNA Integrity Number, and Genomic Quality Number [37].

Furthermore, the trial only incorporated AI into the initial screening of the sample but did not incorporate pathologist feedback for updating predictions, primarily to avoid disturbing the existing clinical workflow of our NGS laboratory. We envision a future pathologist-in-the-loop application, where pathologists may edit macrodissection areas and receive updated predictions in real time. Workflow improvements could also be made to maximize efficiency. For example, a model including imaging-based measures of tissue area and quality combined with clinical data could be used to flag samples up front that may need AI-assistance, while passing samples with high likelihood to succeed in NGS. Future models may also be trained to predict not only DNA yield, but RNA yield and other quality control metrics through the NGS pipeline.

Despite the limitations described, SmartPath accurately predicted tissue quantities needed for adequate DNA yield and provides a viable alternative to manual estimation. SmartPath could be useful in circumstances where access to pathologists is scarce, or for laboratories processing large volumes of tissue. Coupled with a digital slide viewer, such a system can support fully remote pathology review of digitized WSIs, allowing NGS laboratories to widen their range of pathologists to review samples. Integration of SmartPath with automated microdissection systems [15,16,19] could allow for tissue extraction workflows which are almost entirely automated, with a pathologist needed only to approve or modify input parameters, and potentially be economically and clinically beneficial for NGS laboratories.

# Data Availability Statement

Most data generated or analyzed during this study are included in this article and its supplementary information files. Raw data are not made available due to their proprietary nature, but are available from the corresponding author on reasonable request.

# Figure Legends

**Figure 1. SmartPath model pipeline and internal evaluation trial design.**
**A)** SmartPath consists of several models which extract information from H&E-stained whole-slide images and make predictions used to augment pathologist decisions prior to DNA extraction. These models can be grouped into "Feature Generation" (blue boxes) and "DNA Yield Estimation" (green boxes). Arrows point in the direction of data flow between models. The feature generation pipeline receives inputs from pre-trained cell segmentation (U-Net) and tumor segmentation (multi-field-of view ResNet-18) models to generate features used for a DNA Yield Estimator. For samples that are macrodissected, these features are only computed from the macrodissection area by masking the slides, either by a U-Net-based ink detection (when run on archival slides for training) or by post-processing the tumor segmentation model output (when running on new slides during inference). The output of the feature generation pipeline is fed to a regularized linear model to predict the expected DNA yield per slide. During the trial, the predicted DNA yield per slide is used to estimate the total number of slides needed to achieve the following target yields: 100ng, 400ng, 1000ng. These predictions, along with the predicted microdissection area, are output to a UI presented to the pathologist during review.
**B)** Samples either receive Traditional ("Trad") or AI-augmented pathology review ("SmartPath"). In Trad path review a pathologist reviewed a slide under a microscope and estimated the number of slides needed for DNA extraction. Slide scanning was not incorporated into the Trad workflow. In SmartPath pathology review, slide scanning was incorporated into the workflow immediately after tissue was sliced. Slide scanning triggered an upload of the image to the cloud, where it was automatically processed by the SmartPath pipeline described in **A**. At the end of the Trad or Smart review process the pathologist made a review decision to recommend the number of slides scraped as well as the macrodissection area (if needed) for DNA mass extraction. Routine NGS proceeded after extraction. If NGS failed any QC step along the way the sample was re-extracted. Four metrics, marked by circled numbers in the diagram, were used to measure the impact of AI-assistance: 1) the number of slides scraped for extraction, 2) the extracted DNA mass, 3) the number of extraction attempts (referred to as "extraction count"), 4) the time elapsed from $1^{st}$ extraction to $1^{st}$ successful DNA sequencing (referred to as T-seq). All re-extracted samples were reassessed with traditional path review in order to minimize disruption of our existing clinical workflow.

**Figure 2. Scatter plots of predicted vs true DNA mass per slide.**
An unregularized model extremely overfit to the training set **(A)** and failed to generalize to the withheld validation set **(B)**. Predictions from the best performing model (log transform, L1, α=0.01) strongly correlated to both the training set **(C)** and withheld validation set **(D)**. Both axes are log-transformed. Each point represents an extraction attempt and is colored by macrodissection status (Light gray – macrodissected, dark gray – whole-slide dissected). Note that macrodissected slides tend to have higher mass per slide, because they tend to have larger surface area.

**Figure 3. Fraction of samples with DNA mass in target range is significantly increased using AI assistance**
**A)** The fraction of samples reaching the target (100-2000ng), undershoot (<100ng), or overshoot (>2000ng) DNA mass upon first extraction compared between SmartPath and Trad cohorts. Comparisons were then stratified by **B)** tissue size and **C)** extraction quality, where seven samples were excluded due to lack of extraction quality data. To compute statistics, each sample was assigned 1 or 0 whether it was inside or outside the indicated mass range. Error bars are 95% confidence intervals. Significance was determined by a chi-squared test for binary variables performed on the contingency table of counts split by Trad and SmartPath cohorts.

**Figure 4. AI assistance offers more nuanced suggestions of number of slides to scrape for extraction**
**A)** Distribution of N slides scraped plotted for all samples (left), only large tissues (middle), and only small tissues (right). Without AI-assistance, pathologists tended to recommend either 5 or 10 slides for scraping (Trad, orange), but with AI-assistance the distribution was much broader (SmartPath, blue). SmartPath distribution is shifted towards fewer slides for large tissues and more slides for small tissues. **B)** Box plots comparing numbers of slides scraped for SmartPath and Trad cohorts grouped by large and small tissues. P-values were computed from t-tests on log-transformed data assuming unequal variance. White dots - mean. Horizontal black line - median **C)** Truncated violin plots comparing number of sides scraped between SmartPath and Trad cohorts, grouped by large (top) and small (bottom) tissues and by extraction quality. Back boxes - 25% and 75% percentiles. White dots – medians. Samples that did not have recorded numbers of slides scraped were dropped. P-values for Welch's two-sided t-test assuming unequal variance are displayed above each group. Bimodal distributions for the Trad cohort (orange) correspond to 5 and 10 slides.

**Figure 5. Number of extractions needed to reach DNA sequencing.**
**A)** The numbers of extraction attempts needed to reach DNA sequencing are counted for SmartPath and Trad cohorts. Only 4 samples in the total dataset had extraction count = 4, and only one had extraction count = 5. **B)** Mean number of extractions grouped by extraction quality. Error bars are 95% confidence intervals produced with 1000 bootstraps. High-quality samples have no error bars because they are already at optimum, with only one extraction per sample, regardless of cohort. To compute p-values, Poisson distributions were fit to the zeroed distributions (see Section S11) of SmartPath and Trad cohorts and a test was performed to assess if the ratio of the two Poisson rates is statistically different from 1. **C)** Same as **B** but grouped by large and small tissues.

**Figure 6. Mean time to first DNA sequencing of SmartPath cohort relative to Trad cohort.**
**A)** Distribution of time to first DNA sequencing (T-seq) for all samples in the SmartPath and Trad cohorts. **B)** Time from first extraction to first DNA sequencing was measured for SmartPath and Trad cohorts. The plot shows the mean T-seq for SmartPath and Trad cohorts for all samples (left) as well grouped by extraction quality (low, intermediate, high). Error bars are the average of the 5-95% confidence intervals of Trad and SmartPath cohorts. **C)** Same as **B** but grouped by large and small tissues. Significance between the cohort means was assessed by performing t-tests on the log-transformed T-seq.

# Table Legends

**Table 1**
\* Sample age is defined as the delta between time of first extraction attempt and time of sample collection. Because sample age is a continuous variable, a chi-squared test could not be performed. Instead, a t-test was performed on the log-transformed data.
Counts per category are shown for each characteristic grouped by Trad and Smart cohorts, except for characteristics with high cardinality which only show the number of unique categories. These counts define a contingency table for each covariate. A chi-squared test was run on each contingency table to obtain p-values assessing a significant difference between Trad and Smart cohorts. Sample-level data had no data missingness, and in some cases showed significant imbalance, as evidenced by the small chi-squared test p-values.

**Table 2**
\* Indicates P-value significance below 0.05
† AIC - Akaike Information Criterion. The AIC is a fitness parameter that trades off the complexity of a model with how well the model fits the data. It can be interpreted as a measure of model parsimony, where lower value indicates a more parsimonious model. It is a relative measure, and thus can only be compared between models for a given metric.
‡ Covariates are chosen based on significant association in the univariate analysis, hence not every outcome metric is modeled with the same covariates.
☩ Because Pathologist is a categorical variable and thus has p-values for each category, only the most significant p-value is shown.

# Acknowledgements

We thank Timothy Baker for assisting with NGS workflow data acquisition, Dr. Lingdao Sha and Dr. Andrew Kruger for help generating handcrafted imaging features, Dr. Mark Carty for insightful discussions on the statistical analysis of covariates, Dr. Tim Taxter for participating in the validation trial, and Matthew Kase for assistance in reviewing the text and figures. Finally, we are grateful to Erin MCarthy, the Tempus histology lab, and all members of the Tempus NGS lab, who made the internal validation trial possible.

# Author Contributions

B.L.O., R.D.J., A.W., and B.M.M. performed study concept and design; B.L.O., R.J., A.B.T., R.P.J. and M.C.S. performed development of methodology and writing, review and revision of the paper; B.L.O. and I.H. wrote and validated the algorithm code; M.C., C.W., and L.S. built the browser-based UI and coordinated deployment to production; R.D.J. and A.W. supervised the validation trial; R.D.J. and B.M.M. reviewed samples in both SmartPath and Trad arms during the validation trial; B.L.O., A.B.T., and R.P.J. provided acquisition and analysis of the validation trial results. B.L.O., A.B.T., and R.P.J. provided interpretation of data and statistical analysis. All authors read and approved the final paper.

# Funding
This work was supported by Tempus Labs.

**Table 1. Sample-level characteristics of evaluation trial dataset**

|  | Trad (N=233) | SmartPath (N=243) | Chi-sq. or t-test p-value |
|---|---|---|---|
| **Sample age at extraction (days)*** | | | 0.06 |
| Median | 37.57 | 47.58 | |
| Range | 5.05 - 2443.01 | 5.58 - 3205.58 | |
| **Procedure Type** | | | 0.19 |
| Biopsy (unspecified) | 68 | 93 | |
| Needle Biopsy | 53 | 42 | |
| Resection | 112 | 108 | |
| **Dissection** | | | 0.73 |
| Macrodissected | 119 | 130 | |
| Whole slide | 114 | 115 | |
| **Pathologist** | | | < 1e-60 |
| A | 100 | 1 | |
| B | 47 | 205 | |
| C | 36 | 0 | |
| D | 24 | 4 | |
| E | 23 | 1 | |
| F | 3 | 34 | |
| **Extraction day of week** | | | 0.0008 |
| Monday | 23 | 26 | |
| Tuesday | 42 | 17 | |
| Wednesday | 50 | 60 | |
| Thursday | 36 | 37 | |
| Friday | 30 | 58 | |
| Saturday | 45 | 45 | |
| Sunday | 7 | 2 | |
| **Sample characteristics with high cardinality (only showing N unique values)** | | | |
| **Tissue site** | 38 (unique) | 41 (unique) | 0.38 |
| **Extraction tech** | 18 (unique) | 24 (unique) | 0.0001 |

**Table 2. Statistics for covariates fit to trial metrics in samples with small tissue area and low extraction quality (N = 50)**

| Outcome Metric | Univariate GLM (trial cohort only) | | Multivariate GLM (trial cohort + covariates) | | | |
|---|---|---|---|---|---|---|
| | AIC † | p-value | AIC † | Trial Cohort p-value | Covariate ‡ | Covariate p-value |
| undershoot (True if < 100ng) | 54.57 | 0.055 | 55.04 | 0.072 | Day-of-week | 0.22 |
| N slides scraped | 65.52 | 0.075 | 74.11 | 0.14 | Sample age | 0.84 |
| | | | | | Pathologist ✢ | 0.37 |
| Extraction count | 120.21 | 0.049 * | 118.91 | 0.057 | Day-of-week | 0.12 |
| | | | | | Sample age | 0.19 |
| T-seq | 51.90 | 0.026 * | 45.73 | 0.062 | Day-of-week | 0.006* |
| | | | | | Pathologist ✢ | 0.11 |

Figure 1

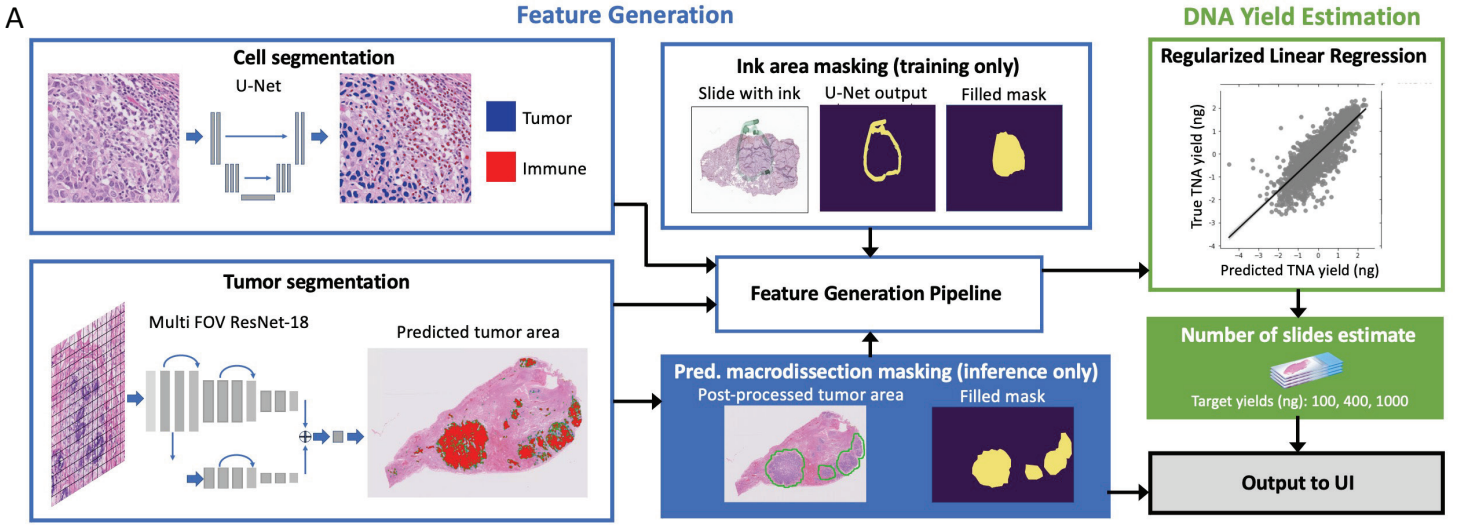

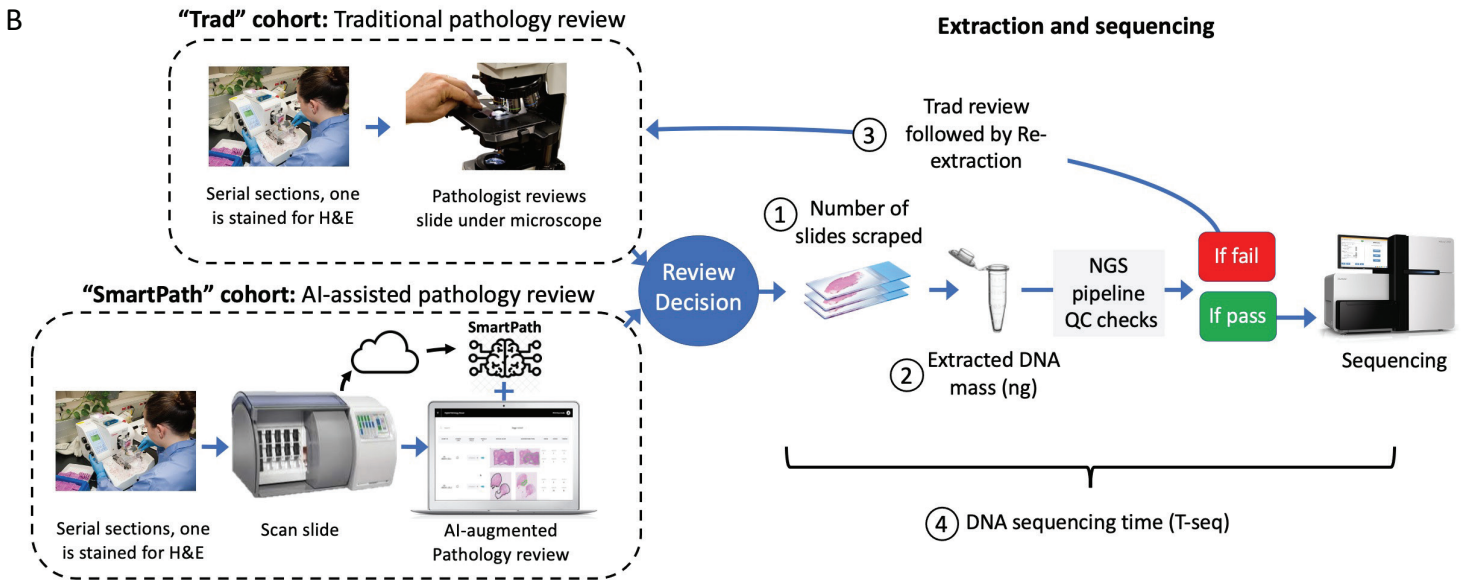

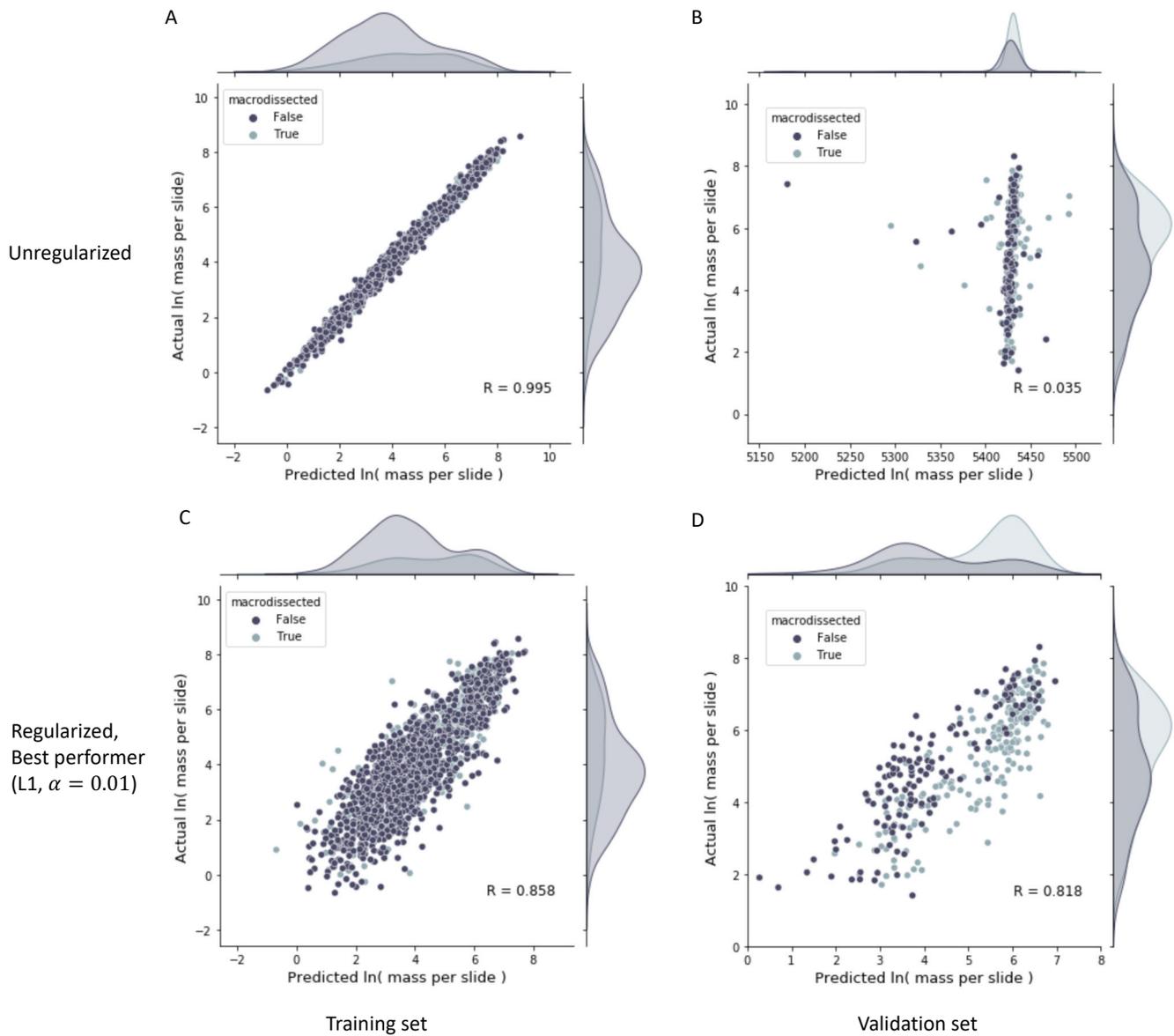

Figure 2. Scatter plots of predicted vs true DNA mass per slide.

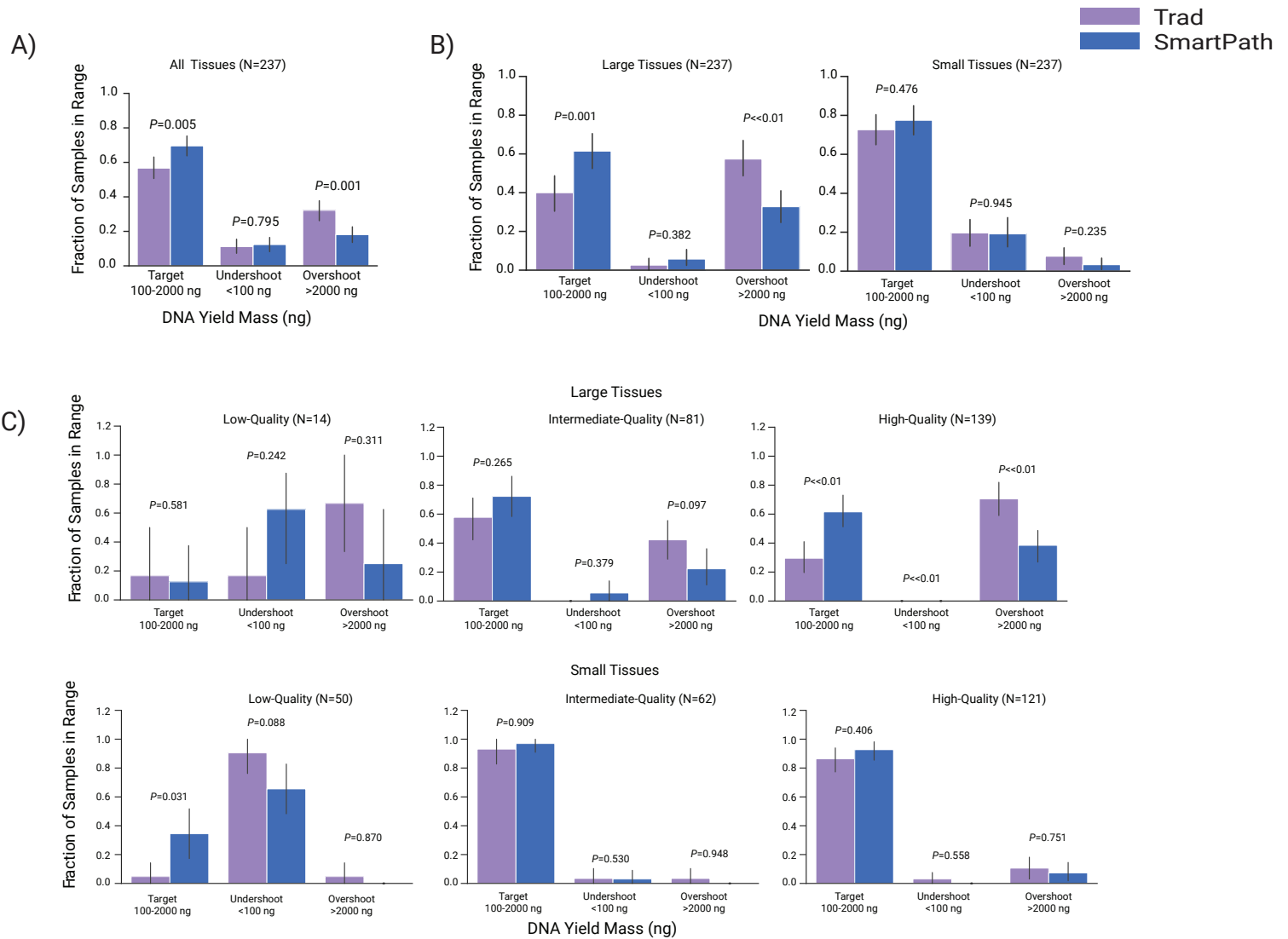

Figure 3. Fraction of samples with DNA mass in target range is significantly increased using AI assistance

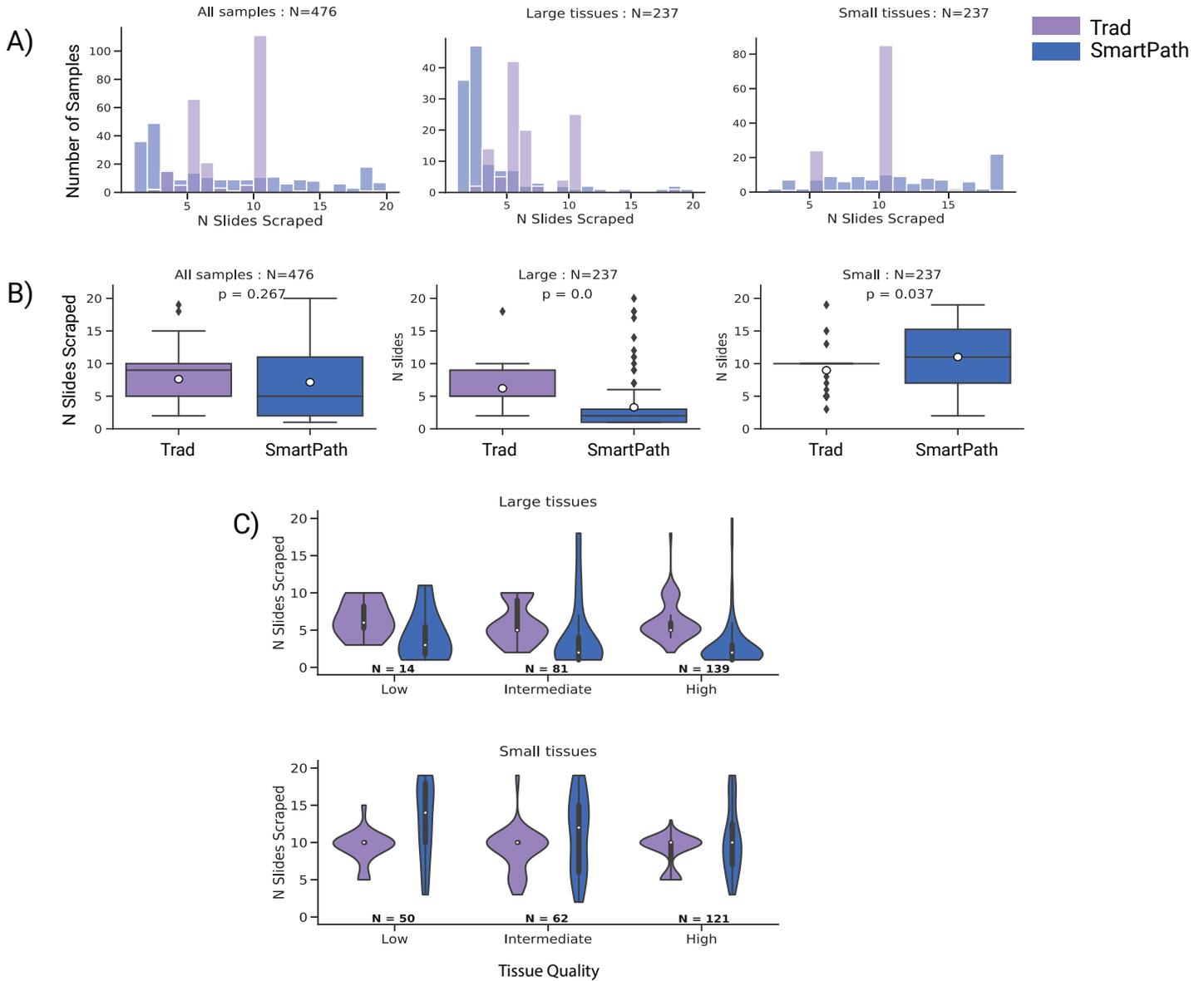

Figure 4. AI assistance offers more nuanced suggestions of number of slides to scrape for extraction

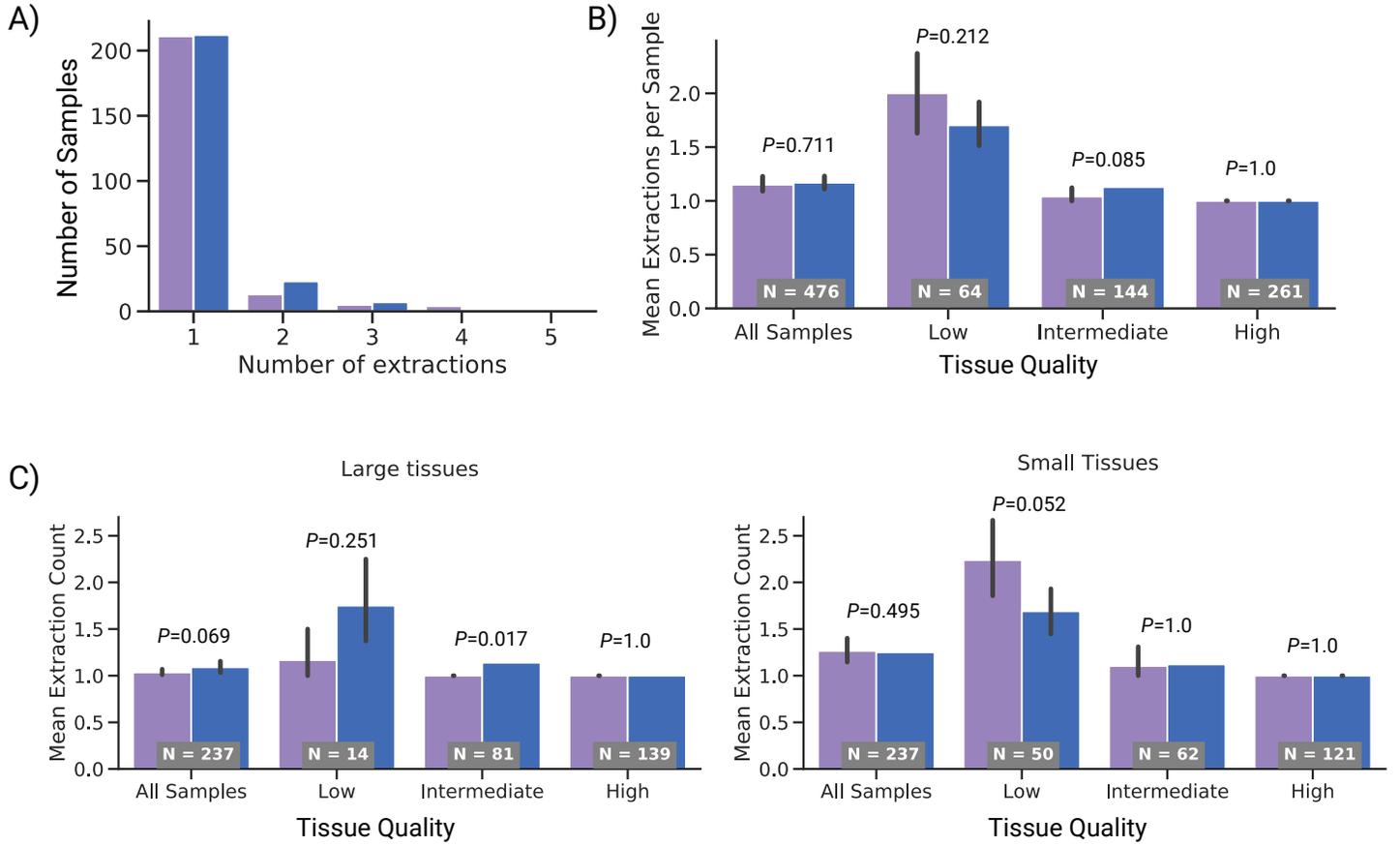

Figure 5. Number of extractions needed to reach DNA sequencing

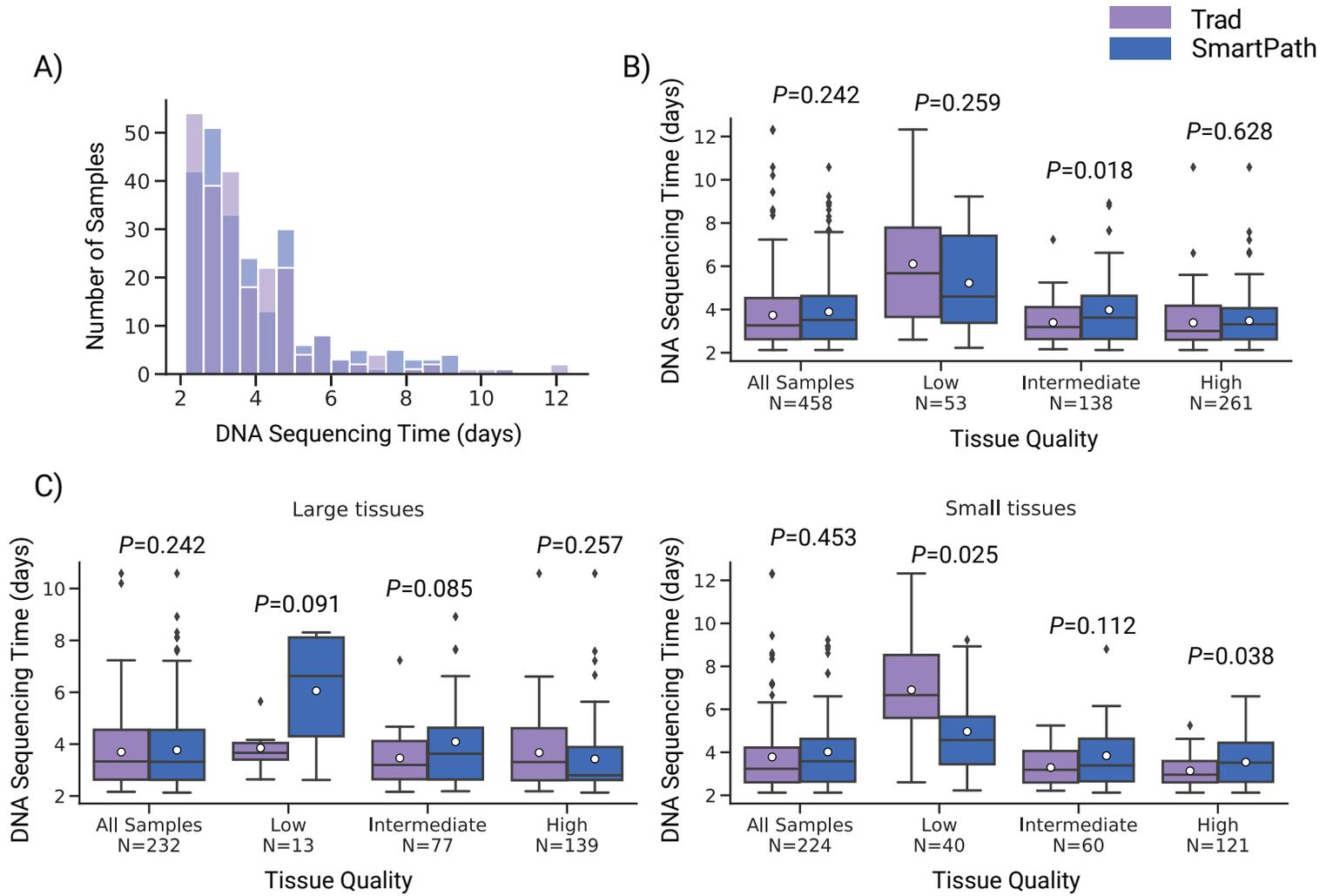

Figure 6. Mean time to first DNA sequencing of SmartPath cohort relative to Trad cohort.

# Supplementary Text

## S1 Tumor and cell segmentation models

*S1.1 Tumor and Immune tissue segmentation*

As part of the feature generation pipeline, we relied on a previously trained fully convolutional network (FCN) architecture with a Res-Net18 backbone for classification of tumor and immune-rich regions from H&E-stained histopathology slides. Details of the architecture are published elsewhere [1], so here we only describe the model training. Three separate models were trained to classify three primary cancers: Colorectal, NSCLC, and Breast. Numbers of slides in each training set were 100 Colorectal, 77 NSCLC, and 100 Breast. Numbers of slides in withheld test sets were 18 Colorectal, 11 NSCLC, and 10 Breast. For ground truth collection the tumor and immune areas were annotated by American Board of Pathology-certified pathologists using the publicly available digital pathology software QuPath [2]. For the test slides we asked pathologists to annotate at least 80% of the tissue area but allowed lower coverage for training slides. Each annotation was converted into grayscale masks and downsampled by 32x. Each pixel of the annotation corresponded to a 466 x 466 tile in the original slide with a stride of 32 pixels at 10x magnification (1 pixel = 1 µm). This tile size was chosen to provide spatial context to the network, while the stride was chosen to increase sampling density and the number of training examples. Only tiles whose center fell within the annotated regions were kept. This procedure produced over one million training tiles and over 500,000 withheld test tiles for each cancer type.

We trained our models on a single NVIDIA V100 GPU by stochastic gradient descent with batch size 100. The FCN ResNet-18 backbone was initialized with pretrained ImageNet weights. Image augmentations (random crop, random rotate, random flip, color jitter) were performed batchwise during training. Rather than generating all tiles and storing them before training, which is costly, we generated the batch on every step of training by reading regions of the WSI that fell within the annotation masks using a custom wrapper around Python OpenSlide [3]. The number of steps in an epoch was set to the total number of tiles divided by the batch size. Batch normalization [4] was also implemented to accelerate training and improve regularization of our network. We used a cross entropy loss function with Adam optimizer and an initial learning rate 1e-3, which was decreased by 0.5 at epochs 3, 5, 7, and 9. Using a train/val split of 80/20 the models were trained with 10 epochs, after which validation loss was observed to no longer decrease. The AUCs on the withheld test set were 0.929 (Tumor) & 0.974 (Immune) for Colorectal, 0.947 (Tumor) & 0.975 (Immune) for NSCLC, 0.966 (Tumor) & 0.998 (Immune) for Breast.

## S1.2 Nucleus segmentation by U-Net

We trained a U-Net [5] architecture written in PyTorch (torchvision 0.2.1) to detect tumor & immune cells in H&E images. We trained 2 separate models; a lymphocyte detector trained on annotations of only lymphocytes and a general nucleus detector trained on annotations of all cell types (mainly lymphocytes, tumor, stroma, and epithelial cells). Annotations were manual outlines of cell nuclei performed in QuPath [2]. The training set consisted of 100 colorectal slides, 50 of which were scanned in house and 50 of which were from the TCGA CRC dataset (using diagnostic slides only). For each slide, 3-5 fields of view (1024 x 1024 at 40x magnification) were annotated. Borders and interiors of annotated cells were labeled as separate classes so as to encourage the model to learn to separate touching cells. Prior to training, field of view images and corresponding masks were downsampled to 20x and enlarged by reflection along each edge. During training these image/mask pairs were randomly cropped to 512x512 and passed through a series of geometric and color augmentations. A range of hyperparameters were explored. Model selection was performed on an 80/20 train/validation split, where the model checkpoint with highest validation DICE score was selected. Final model was trained for 20 epochs using Adam optimizer with learning rate 1e-4 (scheduler reduced by half every 3 epochs) and batch size 16.

## S1.3 Cell-type classification by combining U-Net and FCN ResNet18 model outputs

The U-Net-based cell segmentation models were combined with the FCN ResNet18 based tumor classification model in order to classify individual cells. Cells within a tile were given one of two classes based on the following conditions
- If a cell is detected by the lymphocyte segmentation model then it is classified as a lymphocyte.
- If a cell is detected by the general cell segmentation model but not by the lymphocyte segmentation model, then it is classified as the class of the tile.

For example, if the tile class is "Tumor", then the cell class may be either "lymphocyte" or "tumor cell". If the tile class is "Stroma", then the cell class may be either "lymphocyte" or "stromal cell".

# S2 Description of features output from feature generation pipeline

The feature generation pipeline outputs 3,641 features per slide. These features can be grouped into four general categories, which are described in the following table.

| Feature group | N features | Description |
| --- | --- | --- |
| Cell counts | 5 | <ul><li>Total number of classified cell nuclei</li><li>Number of tumor cell and lymphocyte nuclei</li><li>% tumor cells and % lymphocytes</li></ul> |

| Tumor Shape | 96 | • 8 features describing shape of tumor areas (i.e. area, perimeter, and circularity)<br>• 12 derived summary statistics per feature. |
|---|---|---|
| Cell shape | 660 | • 22 features describing shape of classified cell nuclei (i.e. area, perimeter, and circularity)<br>• 30 derived summary statistics per feature. |
| Cell Texture | 2,700 | • 90 features describing color and texture features of bounding boxes enclosing each detected cell nucleus (i.e mean RGB value, saturation, entropy, Canny edge features)<br>• 30 derived summary statistics per feature. |

The sample size for cell shape and cell texture features was limited to 3,000 randomly selected cells to avoid excessively long computation time for large tissues with 10s of thousands of cells. 1 - 99 percentile clipping was implemented in the feature generation pipeline in order to suppress statistical outliers due to artifactual cell segmentations. All code for the feature generation pipeline is written in Python 3.7.

## S3 Ink detection and macrodissection area estimation during training

Tissues containing a large proportion of non-tumor tissue need to be macrodissected in order to enrich the percentage of tumor cells that are extracted from the tissue by scraping. In traditional pathology review, the macrodissection area is estimated by a pathologist to include an area that contains at least 20% tumor cells, and is hand drawn onto the slide. This area is later manually scraped by a technician.

Approximately 30% of the training samples were macrodissected. For these slides, the computed imaging features must be only from the region within the hand-drawn marker. To do this we first run each slide through a U-Net model previously trained to produce binary masks of the detected ink on the slides (torchvision 0.2.1). Next, we post-process the masks using a custom algorithm (Python OpenCV v 3.4.9.11) to fill the area within the marker. Special care is taken for markings that do not fully enclose an area. Examples of such difficult to process markings, as well as the output of our marker area filling algorithm, are shown Supplementary Figure 16.

The resulting macrodissection area mask is fed to the feature extraction algorithm, to restrict all features within the macrodissection area. This macrodissection area mask, which is generated by filling the area enclosed by hand-drawn ink on a WSI, is only used during feature extraction for model training. During inference a different binary mask, representing the estimated optimal macrodissection area, is used.

## S4 Qualitative validation of microdissection area estimation during inference

As described in the main text, during inference the microdissection area is predicted by post-processing the output of the tumor segmentation. To verify that this procedure produced macrodissection area estimations similar to those hand-drawn by pathologists, the model was applied to 190 slides in the model development validation set previously marked with ink for macrodissection (Supplementary Table 1). Before macrodissection areas were predicted by the model, ink was wiped off the slides to avoid the presence of ink guiding the predictions. The 190 pairs of hand-drawn and predicted areas were compared side-by-side by one of our board-certified pathologists who determined they were close enough in agreement to be useful in practice (Supplementary Figure 17A). It is important to note that the predictions are not meant to replace the pathologist, but only guide them in drawing the macrodissection mask and estimating the numbers of slides to scrape. During the trial, the macrodissection area was predicted for every slide, but the decision whether to use the predicted area and/or number of slides was left to the pathologist's discretion at the time of SmartPath-assisted review. An example of SmartPath-assisted macrodissection workflow in which the pathologist did agree with the prediction is depicted in Supplementary Figure 17B.

## S5 DNA extraction from FFPE slides

To extract DNA from FFPE tissue, serial sections were cut from an FFPE block at a thickness of 4-5 um and affixed to glass slides. A representative slide chosen for H&E staining was baked in a Premier Scientific Slide Warmer and then loaded onto a Tissue Tek Prisma + (Sakura, Torrance, CA) automatic stainer. H&E staining was then reviewed by a board-certified pathologist. If enrichment for tumor cells was needed, the pathologist hand-drew a macrodissection area on the slide, highlighting the tumor area(s) to be scraped. The target tumor area on H&E slides were matched and traced with the unbaked, unstained slides to be scraped.

The unstained slides were carefully scraped with a fresh, sterile razor blade into a labeled 1.5 mL microcentrifuge tube. To deparaffinize the tissue, a series of xylene and ethanol washes and centrifugations were performed where the supernatant was carefully pipetted off to preserve the tissue pellet. The sample tissue pellet was then centrifuged on a vacufuge until all residual ethanol evaporated. Lysis buffer and proteinase K were added to the samples before incubating on thermomixers for two cycles: 1 hour at 56°C with agitation to lyse the cells, and a second 1-hour incubation at 80°C to reverse crosslink the samples. The samples were then loaded onto the automated chemagic 360 instrument to extract total nucleic acid (TNA) using the instrument's chemagen magnetic bead technology. Double-stranded content of the extracted TNA was then quantified by Synergy H1 (Agilent BioTek, Winooski VT) fluorometric quantitation using PicoGreen. The fragmentation quality of each sample was assessed using the Fragment Analyzer (Advanced Analytical Technologies, Ames, IA).

# S6 Details on computing DNA yield per side for ground truth of the model

As described in the main text, the ground truth for fitting the DNA yield predictor was defined differently when only one or multiple extraction attempts were made. Because the digitized H&E slide itself was never extracted, we rely on the assumption that the DNA yield and the model predictions on consecutive slides were constant. To confirm this assumption, we performed a study to measure variability of predicted tumor percent and total cell count per slide. For most slides, this variability was near 5% (Supplementary Figure 1), suggesting that this is a fair approximation. Using this assumption, we computed the ground truth label for each slide by dividing the known DNA yield by the number of slides that were scraped for the extraction.

However, there is an additional consideration. Our lab takes 20 serial sections at initial slicing and stains an intermediate slide with H&E. Therefore, the imaged slide may not always be closest to the sections used for the first extraction attempt. For example, the first extraction may use 5 slides, while the second uses an additional 5 slides, in which case the stained H&E is closest to the slides in the second extraction attempt. For samples that had more than one extraction attempt, we considered the ground truth DNA yield to be the mean of the first two extractions.

# S7 Additional considerations for model inference during the prospective trial

Because the inputs to the DNA Yield prediction model were power transformed, they were required to be strictly positive. This was trivial to ensure during training and validation because the researcher could shift feature values by the minimum of each feature distribution. However, during the prospective trial this was not possible because the model was run one sample at a time and there was no distribution of features available. To ensure strict positivity was met for inference during the trial, we saved values computed from the combined training and validation sets and loaded them during inference. Specifically, we computed the 0.5th and 99.5th percentiles of the distributions for each feature and saved them to an array. These percentile values were hand-picked to fall at the very end of the distributions while excluding extremely large or small outliers. Each feature distribution was shifted by its 0.5th percentile + 1e-3 to ensure strict positivity. The maximum of this shifted distribution was also saved to an array for later normalization.

In rare cases, a feature produced by the feature generation pipeline was found to be not a number (NaN), due to nucleus detection artifacts. Therefore, at the time of inference all NaNs were set to 0 before feature clipping and shifting of means by precomputed values. This ensured that a spurious NaN value did not cause an inference error during the internal trial.

## S8 Details on selection of target yields for number of slides prediction

As described in the main text, SmartPath recommendations were made for three target yields (100 ng, 400 ng, 1000 ng). To select these operating points, we simulated the total predicted DNA yield expected for each sample in the validation set over a range of operating points. The simulated total predicted DNA yield was computed as follows

$$N_{pred} = ceil\left(\frac{Target\ yield}{exp(model_{pred})}\right),$$

$$Y_{pred} = \frac{N_{pred}}{N_{true}} \times Y_{true}$$

where "*Target yield*" is the operating point (in ng), $model_{pred}$ is the predicted log(yield per slide), $N_{pred}$ is the predicted number of slides needed to achieve the target yield, $Y_{true}$ is the true total yield, $N_{true}$ is the true number of slides scraped. Because the model predicts the logarithm of yield per slide, the exponent must be taken to obtain units of yield.

With the predicted total yield ($Y_{pred}$) and the predicted numbers of slides to scrape ($N_{pred}$), we define a Percent Success as the percent of samples satisfying these two constraints:
1) 60 ng < $Y_{pred}$ < 2000 ng
2) $N_{pred}$ < 30 slides

These constraints were determined by our collaborating pathologists to be reasonable definitions of a successful DNA yield. Using the model with optimal parameters trained on the full training set (Supplementary Table 1), we computed the Percent Success of the predictions on the validation set over a range of operating points (Supplementary Figure 4A).

Operating points of 100, 400, and 1000 ng were chosen to roughly span the range in which the estimated Percent Success is higher than the true Percent Success. The highest operating point of 1000 ng exceeds the regions where Percent Success is higher using the model. This is because Percent Success is not the only metric of interest. There is a trade-off between how much yield is achieved and how many slides are scraped, but the costs associated with them are not equal. The costs associated with insufficient yields include the time it takes to perform another extraction (often delayed until the next day) as well as the cost of materials for that extraction, while the cost of scraping too many slides is only the cost of the tissue itself. Therefore, our collaborating pathologists chose to bias the highest operating point towards limiting the number of insufficient yields at the expense of scraping more slides. As can be seen in Supplementary Figure 4B, the estimated percent of samples with insufficient yields is almost zero at an operating point of 1000 ng.

Having multiple operating points is useful in conditions where tissue is limited. For example, the model may recommend 5 slides to achieve >= 100 ng and 20 slides to achieve >= 400 ng, but there are only 7 sections left in the block. If the model only output a single prediction at the 400 ng target, the pathologists would have a harder time estimating if 7 slides was really enough to achieve a yield within 100 – 2000 ng. But because the 100 ng target is present, with 5 slides recommended, the pathologist has more confidence that scraping all 7 slides will give a sufficient DNA yield. In the other extreme when tissue is not limited, the 1000 ng target is a useful choice when the pathologist wants to ensure that sufficient material is extracted to do both DNA and RNA sequencing, which is common in our clinical workflow.

## S9 Estimation of sample size for internal trial

We calculate the sample size needed to measure a difference between the true and predicted Percent Success from the training set (Supplementary Table 1) at chosen significance and power levels. Sample size, $\eta$, is estimated using the formula for two independent samples with binary outcomes [(6)], defined as

$$\eta = 2\left(\frac{Z_{1-\alpha/2} + Z_{1-\beta}}{ES}\right)^2$$

where α is the selected significance level, $Z_{1-\alpha/2}$ is the value from the standard normal distribution 1- α/2 below it, 1- β is the selected power, $Z_{1-\beta}$ is the value from the standard normal distribution holding 1- β below it, and *ES* is the effect size. Effect size is calculated as

$$ES = \frac{|p_1 - p_2|}{\sqrt{p(1-p)}}$$

Where $p_1$ and $p_2$ are the values of the two groups to be compared and *p* is the mean of $p_1$ and $p_2$. Taking the level of significance to be α = 0.01 and the power to be 1- β = 0.80, the values from the standard normal distribution are $Z_{1-\alpha/2}$ = 2.576 and $Z_{1-\beta}$ = 0.84.

We take $p_1$ = 0.67, which is the approximate observed Percent Success for the training set (Supplementary Table 1) and vary $p_2$ over a range of values representing the expected Percent Success that would be achieved using the DNA yield prediction model. The sample sizes, calculated for each value of $p_2$ are presented in the table below.

| p2 value | Effect size | Sample size (per group) |
|---|---|---|
| 0.7 | 0.064583 | 5595 |
| 0.75 | 0.176304 | 751 |
| 0.8 | 0.294562 | 269 |
| 0.85 | 0.421464 | 131 |
| 0.9 | 0.559853 | 74 |
| 0.95 | 0.713738 | 46 |
| 1 | 0.889055 | 30 |

**Sample size estimates for internal trial**

Sample size was computed over a range of p2 values, which represent the Percent Success expected when using the DNA yield prediction model. p2 = 0.70 represents a model with Percent Success = 70%, while p2 = 100 represents a model with highest possible Percent Success. Computation assumes a significance level of 0.01 and a power of 0.80.

Taking p2 = 0.80 - 0.85 as a reasonable range to expect for model performance, the sample size of each group should be between 269 and 131. Based on this estimate we choose a sample size of 250 for the traditional and AI-assisted groups of the internal trial.

# S10 Model deployment for internal trial

For the internal trial, we orchestrated a series of automatically triggered model inference steps run on AWS EC2 instances using AWS Lambda (Supplementary Figure 5). Slide scanning automatically triggered tumor and cell segmentation inference, which were run on a p3.2xlarge instance (one Tesla V100 GPU). This was followed by macrodissection area prediction, feature generation, and number of slides prediction, all of which were run on a m5.8xlarge instance (CPU only). The outputs were automatically uploaded to a browser-based UI built by our engineering team. During pathology review, the pathologists indicated their chosen target yield into the UI. This UI was linked to a pathology viewer application, which allowed the pathologist to view the WSI at high resolution if needed. The model predictions and user inputs were uploaded to an SQL database. Data was collected over a period of 5 months in the latter half of 2020. During the initial 2 weeks of the trial, bugs with the deployment pipeline were encountered and fixed, after which the trial ran with minimal support from engineering. All IDs in the database were de-identified with a custom script that assigns random IDs for analysis.

# S11 Selection of appropriate GLMs for modeling outcome metrics

For each outcome metric, we created GLMs with three fixed independent variables, trial cohort, extraction quality, and tissue area. To measure goodness-of-fit we used the R package DHARMa [(7)] to perform three statistical tests on the simulated vs expected residuals for a given distribution family (Q-Q plot) (Supplementary Figure 8). The final GLMs for each metric are described below.

GLM for DNA mass undershoot: This metric is a binary boolean, which is best modeled by a binomial GLM. Transformation of predictors is unnecessary to obtain a high-quality fit (Supplementary Figure 8A).

GLM for N slides: Exploration of several distributions found that a Gaussian family GLM fit the data well. When coupled with a log transform of N slides, the GLM passed all three residual diagnostic tests (Supplementary Figure 8B).

GLM for extraction count: Because this is count data, a Poisson family distribution was explored. However, Poisson distributions assume a minimum of zero, but the extraction count distribution has a minimum of 1. When diagnostic tests were computed using the raw data, the goodness-of-fit was poor, but when performed on the extraction count minus 1 (referred to as the "zeroed" distribution), the GLM passed all three tests (Supplementary Figure 8C).

GLM for T-seq: A Gamma family with log transformation of the data was fit (Supplementary Figure 8D). This model failed the K-S test, suggesting a deviation of the predictive distribution from the true distribution, though it passed the other two statistical tests.

## S12 Details on encoding of covariates

Trial cohort, indicating whether a sample was in the SmartPath or Trad cohorts, was dummy encoded by the R glm function as a binary indicator (0 - SmartPath, 1 - Trad).

Extraction quality has three categories: low, intermediate, and high, as defined earlier. We chose to numerically encode these values as ordinal variables (0, 1, 2) to simplify the presentation of GLM results. As a sanity check, GLMs were also fit using categorical versions of extraction quality, but the interpretation of the results remained the same.

Extraction day-of-week was encoded numerically from Monday to Sunday as 0-6. Normally the day-of-week is thought of as a category, where each day is independent. However, lab operations did not function every day equally, due to fewer operations during the weekend and accumulation of re-extractions from day to day. In fact, this causes an almost linear increase in T-seq from Monday to Sunday (see Supplementary Figure 13C). Such weekend effects are common in workplaces and were likely compounded by the fact that this trial was run during the lock-down phase of the 2020 COVID-19 epidemic. As a sanity check, GLMs were also fit using categorical versions of extraction day-of week, but the interpretation of the results remained the same.

Sample age distribution had a very large tail, with most samples having ages less than a month but some being over 10 years old (Table 1). Therefore, we took the logarithm of sample age to bring it closer to a linear scale.

Procedure type, pathologist, and extraction tech group are categorical by nature and were dummy encoded by the R glm function. To eliminate correlations between dummy encoded features, one category from each feature was dropped. The dropped categories are Biopsy, Pathologist F, and Tech Group 1 for Procedure Type, Pathologist, and Extraction Tech Group.

## S13 Construction of Univariate and Multivariate GLMs

GLMs are fit using the glm function in R, which takes the distribution family and formula strings as inputs. The formula string for univariate GLMs has the form

$$M \sim covariate$$

, and for multivariate GLMs has the form

$$M \sim trial\_cohort + covariate(s)$$

, where $D$ is one of the outcome metrics. For multivariate models we kept only those covariates that were found to be significantly associated with outcome metrics in univariate GLMs.

Univariate GLMs were initially fit using the sample-level (Table 1) and patient-level characteristics (Supplementary Table 4) as independent variables, but significant effects were not found for any of the patient-level characteristics. Therefore, the univariate GLM coefficient tables presented in Supplementary Tables 5 – 8 are only for sample-level characteristics. Significant effects were also not found for two sample-level characteristics, microdissection status and tissue site, which are not presented in the GLM coefficient tables.

The reported statistics of the GLM fits include the Akaike Information Criterion (AIC) and Null/Residual Deviance. The Akaike Information Criterion (AIC) is an information-theoretic measure that balances model fit with model complexity. Lower AIC indicates a more parsimonious model, while higher AIC indicates possibly too many variables. By measuring the degree of model complexity, the AIC for GLMs plays a similar role to the adjusted $R^2$ of linear models. The Null Deviance is the deviance of the null model just using the intercept, and the Residual Deviance is the deviance after the model has been fit. A favorable model is therefore one that has a low Residual Deviance (relative to Null Deviance), while maintaining a low AIC.

Analysis of covariance was accomplished by comparing the fits of univariate GLMs using only trial cohort to multivariate GLMs, including trial cohort and covariates, which determined if the main effect of the trial cohort remained significant after being adjusted for covariates.

# Supplementary References

# Supplementary Tables

**Supplementary Table 1. Training and validation set characteristics**

|  | Training set N = 1,605 | Validation set N = 332 |
|---|---|---|
| **Cancer type** | | |
| Colorectal | 470 | 332 |
| Breast | 460 | 0 |
| Lung | 675 | 0 |
| **Gender** | | |
| Male | 563 | 175 |
| Female | 877 | 157 |
| Missing | 165 | 0 |
| **Race** | | |
| White | 693 | 146 |
| African American | 113 | 36 |
| Asian | 37 | 6 |
| Native American | 5 | 0 |
| Missing | 757 | 144 |
| **Procedure Type** | | |
| Biopsy (unspecified) | 554 | 101 |
| Needle Biopsy | 548 | 67 |
| Resection | 442 | 164 |
| Missing | 61 | 0 |
| **AJCC Stage** | | |
| IA/IB | 17 | 2 |
| IIA/IIB/IIC | 61 | 15 |
| IIIA/IIIB/IIIC | 95 | 30 |
| IVA/IVB/IVC | 1065 | 232 |
| Missing | 367 | 53 |

**Grade**

| | | |
|---|---|---|
| Grade 1 (well differentiated) | 61 | 24 |
| Grade 2 (moderately differentiated) | 441 | 199 |
| Grade 3 (poorly differentiated) | 474 | 72 |
| Grade 4 (undifferentiated) | 4 | 0 |
| Missing | 624 | 37 |

**Histology**

| | | |
|---|---|---|
| Adenocarcinoma | 903 | 321 |
| Carcinoma, other | 464 | 10 |
| Malignant neoplasm | 54 | 0 |
| Neuroendocrine tumor | 14 | 1 |
| Missing | 170 | 0 |

**Dissection***

| | | |
|---|---|---|
| Macrodissected | 449 | 190 |
| Whole slide | 1156 | 142 |

* The 190 macrodissected validation set slides have two digitized versions, the first scanned with the hand-drawn ink present, and the second scanned after the ink was wiped off.

**Supplementary Table 2. Parameter exploration for DNA Yield prediction model**

| Power Transform | Regularization | Regularization strength | R (train) | R (validation) |
|---|---|---|---|---|
| Natural log | L1 | 0.001 | 0.915 | 0.593 |
| | | 0.01 | 0.858 | **0.818** |
| | | 0.1 | 0.836 | 0.810 |
| | | 1 | 0.753 | 0.792 |
| | L2 | 1 | 0.949 | 0.429 |
| | | 10 | 0.921 | 0.593 |
| | | 1000 | 0.857 | 0.805 |

|  |  | 10000 | 0.803 | 0.767 |
|---|---|---|---|---|
| Box-Cox* | L1 | 0.001 | 0.961 | 0.614 |
|  |  | 0.01 | 0.893 | 0.792 |
|  |  | 0.1 | 0.833 | 0.817 |
|  |  | 1 | 0.0 | 0.0 |
|  | L2 | 1 | 0.998 | -0.307 |
|  |  | 10 | 0.984 | 0.155 |
|  |  | 1000 | 0.911 | 0.717 |
|  |  | 10000 | 0.845 | 0.722 |

* For Box-Cox transforms the same λ parameter used for transforming the training distribution was used to transform the validation set to prevent data leakage. The combination with highest correlation coefficient on validation set (R = 0.818, bold) is a log transform with an L1 regularization strength of 0.01.

**Supplementary Table 3. Acceptance criteria for Smart Path Trial enrollment**

| **Acceptance Criteria** |
|---|
| ● Colonic carcinomas<br>● Colorectal carcinomas<br>● Metastatic carcinoma, c/w colorectal primary<br>● Rectal carcinomas<br>● Appendiceal carcinomas<br>● Mucinous carcinomas, determined to be colorectal primary<br>● Sample is part of a Multi part order (Both H&Es will be scanned)<br>● Small intestine primary |

| Rejection Criteria |
| --- |
| <ul><li>Surgical Procedure is FNA or Fluid aspirate</li><li>Sample was sent to serve as additional tissue for NGS or IHC</li><li>Neuroendocrine carcinoma of the colon</li><li>Diagnosis or Cohort, requires additional confirmation from Pathologist</li><li>Samples with diagnosis of c/w, favor, or suggestive of GI primary</li><li>Samples that may require a re-stain</li><li>Decalcified samples</li><li>Samples that require additional follow up</li></ul> |

**Supplementary Table 4. Patient-level characteristics of evaluation trial dataset**

| | Trad (N=233) | SmartPath (N=243) | Chi-sq. p-value |
| --- | --- | --- | --- |
| **Gender** | | | 0.32 |
| Male | 101 | 123 | |
| Female | 92 | 84 | |
| Missing | 40 | 36 | |
| **Race** | | | 0.84 |
| White | 95 | 94 | |
| African American | 14 | 15 | |
| Asian | 4 | 4 | |
| Native American | 1 | 0 | |
| Missing | 119 | 130 | |
| **Ethnicity** | | | 0.68 |
| Not Hispanic or Latino | 54 | 52 | |
| Hispanic or Latino | 7 | 12 | |
| Missing | 172 | 179 | |
| **AJCC Stage** | | | 0.53 |
| IA/IB | 1 | 1 | |
| IIA/IIB/IIC | 2 | 6 | |
| IIIA/IIIB/IIIC | 12 | 18 | |
| IVA/IVB/IVC | 169 | 165 | |
| Missing | 49 | 53 | |
| **Grade** | | | 0.51 |
| Grade 1 (well differentiated) | 11 | 13 | |
| Grade 2 (moderately differentiated) | 111 | 107 | |
| Grade 3 (poorly differentiated) | 35 | 48 | |
| Grade 4 (undifferentiated) | 0 | 1 | |
| Missing | 76 | 74 | |
| **Histology** | | | 0.45 |
| Adenocarcinoma | 191 | 198 | |
| Signet ring / Goblet cell carcinoma | 1 | 5 | |
| Carcinoma, no subtype | 41 | 40 | |
| Missing | 0 | 0 | |

Counts per category are shown for each characteristic grouped by Trad and SmartPath cohorts. These counts define a contingency table for each covariate. A Chi-squared test was run on each contingency table to obtain p-values assessing a significant difference between Trad and Smart cohorts. For some of the patient-level characteristics there is high missingness, but the overall number of samples between Trad and SmartPath cohort are similar for each characteristic.

**Supplementary Table 5.** Univariate GLM Coefficients of Binomial model for undershoot boolean, fitting a separate model for each covariate.

| | Estimate | Std. Error | z value | Pr(>|z|) | | AIC | Resid. Dev. |
|---|---|---|---|---|---|---|---|
| Numerical & binary variables | | | | | | | |
| trial_cohort | -0.1269 | 0.3001 | -0.423 | 0.672 | | 322.15 | 318.15 |
| extraction_quality | -3.6095 | 0.4196 | -8.603 | < 2e-16 | *** | 149 | 145 |
| tissue_area | 1.8353 | 0.398 | 4.611 | 4.01E-06 | *** | 293.73 | 289.73 |
| extraction_dayofweek | 0.28607 | 0.09727 | 2.941 | 0.00327 | ** | 313.12 | 309.12 |

| | | | | | | |
|---|---|---|---|---|---|---|
| log_sample_age_at_extraction | 0.01365 | 0.09672 | 0.141 | 0.888 | 322.31 | 318.31 |

| Categorical variables | | | | | | |
|---|---|---|---|---|---|---|
| Procedure(Core needle biopsy) | 1.4477 | 0.3515 | 4.119 | 3.80E-05 *** | 276.22 | 270.22 |
| Procedure(Surgical resection) | -1.3053 | 0.495 | -2.637 | 0.00837 ** | | |
| Pathologist A | 0.04049 | 0.3772 | 0.107 | 0.915 | 327 | 315 |
| Pathologist B | -1.33207 | 1.06502 | -1.251 | 0.211 | | |
| Pathologist C | 0.31015 | 0.62744 | 0.494 | 0.621 | | |
| Pathologist D | 0.156 | 0.69496 | 0.224 | 0.822 | | |
| Pathologist E | -0.29598 | 0.68239 | -0.434 | 0.664 | | |
| Tech Group 2 | 1.7014 | 0.8494 | 2.003 | 0.04517 * | 322.54 | 272.54 |
| Tech Group 3 | -16.0538 | 6522.6386 | -0.002 | 0.99804 | | |
| Tech Group 4 | 0.7206 | 1.2357 | 0.583 | 0.55983 | | |
| Tech Group 5 | -16.0538 | 6522.6386 | -0.002 | 0.99804 | | |
| Tech Group 6 | -0.6232 | 0.9392 | -0.664 | 0.50699 | | |
| Tech Group 7 | -16.0538 | 3765.8472 | -0.004 | 0.9966 | | |
| Tech Group 8 | 21.0784 | 6522.6386 | 0.003 | 0.99742 | | |
| Tech Group 9 | -16.0538 | 2917.0127 | -0.006 | 0.99561 | | |
| Tech Group 10 | -16.0538 | 4612.202 | -0.003 | 0.99722 | | |
| Tech Group 11 | -16.0538 | 6522.6386 | -0.002 | 0.99804 | | |
| Tech Group 12 | 0.4082 | 0.7647 | 0.534 | 0.5935 | | |
| Tech Group 13 | -16.0538 | 6522.6386 | -0.002 | 0.99804 | | |
| Tech Group 14 | 0.3657 | 0.696 | 0.525 | 0.59927 | | |
| Tech Group 15 | 2.33 | 0.8527 | 2.733 | 0.00628 ** | | |
| Tech Group 16 | -16.0538 | 2917.0127 | -0.006 | 0.99561 | | |
| Tech Group 17 | -16.0538 | 1423.3566 | -0.011 | 0.991 | | |
| Tech Group 18 | 0.4842 | 0.766 | 0.632 | 0.52732 | | |
| Tech Group 19 | -16.0538 | 2062.6395 | -0.008 | 0.99379 | | |
| Tech Group 20 | 0.9029 | 1.2491 | 0.723 | 0.46981 | | |
| Tech Group 21 | 0.4754 | 0.8586 | 0.554 | 0.57977 | | |
| Tech Group 22 | 1.8192 | 1.3639 | 1.334 | 0.18229 | | |
| Tech Group 23 | -0.3209 | 0.8446 | -0.38 | 0.70397 | | |
| Tech Group 24 | 0.0274 | 0.9497 | 0.029 | 0.97699 | | |
| Tech Group 25 | 2.7355 | 0.9002 | 3.039 | 0.00238 | ** | |

Null deviance = 318.33
Signif. codes:  '***' 0.001 '**' 0.01 '*' 0.05 '.' 0.1

**Supplementary Table 6.** Univariate GLM Coefficients of Gaussian model for ln (N slides scraped), fitting a separate model for each covariate.

| | Estimate | Std. Error | z value | Pr(>|z|) | | AIC | Resid. Dev. |
|---|---|---|---|---|---|---|---|
| Numerical & binary variables | | | | | | | |
| trial_cohort | 0.39286 | 0.07073 | 5.554 | 4.69E-08 | *** | 1084.5 | 273.78 |
| extraction_quality | -0.20225 | 0.04992 | -4.051 | 5.96E-05 | *** | 1098.3 | 281.95 |
| tissue_area | 0.95428 | 0.05814 | 16.14 | <2e-16 | *** | 900.91 | 185.09 |
| extraction_dayofweek | -0.0219 | 0.02217 | -0.988 | 0.324 | | 1113.5 | 291.25 |
| log_sample_age_at_extraction | -0.08753 | 0.02336 | -4.336 | 2.01E-04 | *** | 1100.6 | 283.34 |
| Categorical variables | | | | | | | |
| Procedure(Core needle biopsy) | 0.24472 | 0.08587 | 2.85 | 0.00457 | ** | 945.09 | 202.51 |
| Procedure(Surgical resection) | -0.76665 | 0.06894 | -11.12 | < 2e-16 | *** | | |
| Pathologist A | 0.44434   | 0.19342  | 2.297  | 0.02205 | *   | 1090.8 | 272.81 |
| Pathologist B | 0.59414   | 0.18649  | 3.186  | 0.00154 | **  |        |        |
| Pathologist C | 0.01081   | 0.20228  | 0.053  | 0.95741 |     |        |        |
| Pathologist D | -0.08886  | 0.13691  | -0.649 | 0.51662 |     |        |        |
| Pathologist E | 0.10747   | 0.149    | 0.721  | 0.4711  |     |        |        |
| Tech Group 2  | 0.31061   | 0.247829 | 1.253  | 0.2107  |     | 1119.7 | 267.56 |
| Tech Group 3  | -1.603584 | 0.78592  | -2.04  | 0.0419  | *   |        |        |
| Tech Group 4  | -0.721614 | 0.318043 | -2.269 | 0.0238  | *   |        |        |
| Tech Group 5  | 0.794312  | 0.78592  | 1.011  | 0.3127  |     |        |        |
| Tech Group 6  | 0.131215  | 0.166191 | 0.79   | 0.4302  |     |        |        |
| Tech Group 7  | -1.067104 | 0.464686 | -2.296 | 0.0221  | *   |        |        |
| Tech Group 8  | 0.005854  | 0.78592  | 0.007  | 0.9941  |     |        |        |
| Tech Group 9  | -0.20664  | 0.36822  | -0.561 | 0.575   |     |        |        |
| Tech Group 10 | 0.491244  | 0.562465 | 0.873  | 0.3829  |     |        |        |
| Tech Group 11 | 0.188176  | 0.78592  | 0.239  | 0.8109  |     |        |        |
| Tech Group 12 | 0.131725  | 0.167825 | 0.785  | 0.4329  |     |        |        |
| Tech Group 13 | -0.910437 | 0.78592  | -1.158 | 0.2473  |     |        |        |
| Tech Group 14 | 0.13077   | 0.148567 | 0.88   | 0.3792  |     |        |        |
| Tech Group 15 | 0.677834  | 0.264286 | 2.565  | 0.0107  | *   |        |        |
| Tech Group 16 | 0.05682   | 0.36822  | 0.154  | 0.8774  |     |        |        |
| Tech Group 17 | 0.237428  | 0.20919  | 1.135  | 0.257   |     |        |        |
| Tech Group 18 | 0.266008  | 0.170526 | 1.56   | 0.1195  |     |        |        |
| Tech Group 19 | 0.158634  | 0.274455 | 0.578  | 0.5636  |     |        |        |
| Tech Group 20 | 0.251938  | 0.339852 | 0.741  | 0.4589  |     |        |        |

| | Estimate | Std. Error | z value | Pr(>|z|) | |
|---|---|---|---|---|---|
| Tech Group 21 | 0.06212 | 0.195556 | 0.318 | 0.7509 | |
| Tech Group 22 | 0.667565 | 0.464686 | 1.437 | 0.1515 | |
| Tech Group 23 | 0.062104 | 0.16194 | 0.384 | 0.7015 | |
| Tech Group 24 | 0.457635 | 0.195556 | 2.34 | 0.0197 | * |
| Tech Group 25 | 0.126022 | 0.286393 | 0.44 | 0.6601 | |

Null deviance = 291.86
Signif. codes:  '***' 0.001 '**' 0.01 '*' 0.05 '.' 0.1

**Supplementary Table 7.** Univariate GLM Coefficients of Poisson model for zeroed extraction count, fitting a separate model for each covariate.

| | Estimate | Std. Error | z value | Pr(>|z|) | | AIC | Resid. Dev. |
|---|---|---|---|---|---|---|---|
| **Numerical & binary variables** | | | | | | | |
| trial_cohort | -0.1072 | 0.2488 | -0.431 | 0.666 | | 420.31 | 313.59 |
| extraction_quality | -2.5663 | 0.2840 | -9.038 | < 2e-16 | *** | 246.53 | 246.53 |
| tissue_area_category | 1.4982 | 0.3197 | 4.686 | 2.78E-06 | *** | 392.04 | 285.32 |
| extraction_dayofweek | 0.28079 | 0.08082 | 3.474 | 0.000512 | *** | 407.63 | 300.91 |
| log_sample_age_at_extraction | 0.19225 | 0.07622 | 2.522 | 0.0117 | * | 414.3 | 307.58 |
| **Categorical variables** | | | | | | | |
| Procedure(Core needle biopsy) | 0.89399 | 0.27696 | 3.228 | 0.00125 | ** | 386.90 | 278.18 |
| Procedure(Surgical resection) | -1.0104 | 0.3693 | -2.736 | 0.00621 | ** | | |
| Pathologist A | 1.9128 | 1.0260 | 1.864 | 0.062265 | . | 417.12 | 302.40 |
| Pathologist B | 1.6536 | 1.0138 | 1.631 | 0.102866 | | | |
| Pathologist C | 0.8109 | 1.2247 | 0.662 | 0.507892 | | | |
| Pathologist D | 2.0431 | 1.0801 | 1.892 | 0.058555 | . | | |
| Pathologist E | 0.4055 | 1.4142 | 0.287 | 0.774336 | | | |
| --- | --- | --- | --- | --- | --- | --- | --- |
| Tech Group 2 | 2.10476 | 0.677 | 3.109 | 0.001878 | *** | 416.35 | 263.63 |
| Tech Group 3 | -15.71232 | 5717.53217 | -0.003 | 0.997807 | | | |
| Tech Group 4 | 0.64436 | 1.1547 | 0.558 | 0.576824 | | | |
| Tech Group 5 | -15.71232 | 5717.53217 | -0.003 | 0.997807 | | | |
| Tech Group 6 | -0.58779 | 0.91287 | -0.644 | 0.519648 | | | |
| Tech Group 7 | -15.71232 | 3301.01877 | -0.005 | 0.996202 | | | |
| Tech Group 8 | 3.28341 | 0.91287 | 3.597 | 0.000322 | *** | | |
| Tech Group 9 | -15.71232 | 2556.95817 | -0.006 | 0.995097 | | | |
| Tech Group 10 | -15.71232 | 4042.90579 | -0.004 | 0.996899 | | | |
| Tech Group 11 | -15.71232 | 5717.53216 | -0.003 | 0.997807 | | | |
| Tech Group 12 | 0.70754 | 0.69007 | 1.025 | 0.305214 | | | |
| Tech Group 13 | -15.71232 | 5717.53217 | -0.003 | 0.997807 | | | |
| Tech Group 14 | 0.77498 | 0.63621 | 1.218 | 0.223179 | | | |
| Tech Group 15 | 1.80181 | 0.7303 | 2.467 | 0.013616 | * | | |
| Tech Group 16 | -15.71232 | 2556.95817 | -0.006 | 0.995097 | | | |
| Tech Group 17 | -15.71232 | 1247.66793 | -0.013 | 0.989952 | | | |
| Tech Group 18 | 0.4385 | 0.7303 | 0.6 | 0.548208 | | | |
| Tech Group 19 | -15.71232 | 1808.04251 | -0.009 | 0.993066 | | | |
| Tech Group 20 | 0.79851 | 1.1547 | 0.692 | 0.489234 | | | |
| Tech Group 21 | 0.71846 | 0.76376 | 0.941 | 0.346863 | | | |
| Tech Group 22 | 1.49165 | 1.1547 | 1.292 | 0.196423 | | | |
| Tech Group 23 | -0.01242 | 0.76376 | -0.016 | 0.987023 | | | |

| | | | | | | |
|---|---|---|---|---|---|---|
| Tech Group 24 | 0.71846 | 0.76376 | 0.941 | 0.346863 | | |
| Tech Group 25 | 1.77934 | 0.76376 | 2.33 | 0.019822 | * | |

Null deviance = 313.77
Signif. codes:  '***' 0.001 '**' 0.01 '*' 0.05 '.' 0.1

**Supplementary Table 8.** Univariate GLM Coefficients of Gamma model for ln( T-seq ), fitting a separate model for each covariate.

| | Estimate | Std. Error | t value | Pr(>\|t\|) | | AIC | Resid. Dev. |
|---|---|---|---|---|---|---|---|
| Numerical & binary variables | | | | | | | |
| trial_cohort | 0.02746 | 0.02143 | 1.281 | 0.201 | | 300.43 | 34.395 |
| extraction_quality | 0.10412 | 0.01315 | 7.918 | 1.9e-14 | *** | 243.55 | 30.373 |
| tissue_area_category | -0.01838 | 0.02142 | -0.858 | 0.391 | | 301.41 | 34.468 |
| extraction_dayofweek | -0.084070 | 0.005315 | -15.82 | <2e-16 | *** | 75.623 | 21.020 |
| log_sample_age_at_extraction | -0.004734 | 0.007108 | -0.666 | 0.506 | | 301.73 | 34.492 |
| Categorical variables | | | | | | | |
| Procedure(Core needle biopsy) | -0.060537 | 0.02956 | -2.048 | 0.0412 | * | 298.92 | 33.983 |
| Procedure(Surgical resection) | -0.01165 | 0.02438 | -0.478 | 0.6332 | | | |
| Pathologist A | -0.06546 | 0.04804 | -1.363 | 0.1736 | | 303.99 | 34.063 |
| Pathologist B | -0.08524 | 0.04431 | -1.924 | 0.0550 | . | | |
| Pathologist C | -0.05121 | 0.05924 | -0.864 | 0.3878 | | | |
| Pathologist D | -0.05389 | 0.06070 | -0.888 | 0.3751 | | | |
| Pathologist E | -0.12259 | 0.06036 | -2.031 | 0.0428 | * | | |
| Tech Group 2 | -0.1329 | 0.05863 | -2.267 | 0.0239 | * | 243.62 | 27.348 |

| | | | | | |
|---|---|---|---|---|---|
| Tech Group 3 | 0.44882 | 0.31071 | 1.445 | 0.14933 | |
| Tech Group 4 | 0.13081 | 0.09094 | 1.438 | 0.15108 | |
| Tech Group 5 | 0.03241 | 0.20313 | 0.16 | 0.8733 | |
| Tech Group 6 | 0.18981 | 0.04654 | 4.078 | 5.42E-05 | *** |
| Tech Group 7 | -0.04125 | 0.10928 | -0.377 | 0.706 | |
| Tech Group 8 | -0.26105 | 0.12815 | -2.037 | 0.04226 | * |
| Tech Group 9 | 0.12242 | 0.10489 | 1.167 | 0.24383 | |
| Tech Group 10 | 0.41155 | 0.21399 | 1.923 | 0.05512 | . |
| Tech Group 11 | 0.305 | 0.27348 | 1.115 | 0.26536 | |
| Tech Group 12 | 0.04296 | 0.04348 | 0.988 | 0.32369 | |
| Tech Group 13 | -0.08832 | 0.17213 | -0.513 | 0.60815 | |
| Tech Group 14 | 0.02212 | 0.03778 | 0.586 | 0.55846 | |
| Tech Group 15 | -0.06408 | 0.06616 | -0.968 | 0.33334 | |
| Tech Group 16 | 0.14831 | 0.10778 | 1.376 | 0.16952 | |
| Tech Group 17 | 0.36259 | 0.06974 | 5.199 | 3.11E-07 | *** |
| Tech Group 18 | -0.03036 | 0.04194 | -0.724 | 0.46952 | |
| Tech Group 19 | 0.29448 | 0.09047 | 3.255 | 0.00122 | ** |
| Tech Group 20 | 0.09019 | 0.10131 | 0.89 | 0.37385 | |
| Tech Group 21 | 0.06479 | 0.0513 | 1.263 | 0.20735 | |
| Tech Group 22 | 0.14673 | 0.13662 | 1.074 | 0.28342 | |
| Tech Group 23 | 0.09199 | 0.04279 | 2.15 | 0.03215 | * |
| Tech Group 24 | 0.06019 | 0.05112 | 1.177 | 0.23966 | |
| Tech Group 25 | -0.05456 | 0.07404 | -0.737 | 0.46164 | |

Null deviance = 34.529

Signif. codes:  '***' 0.001 '**' 0.01 '*' 0.05 '.' 0.1

# Supplementary Figures

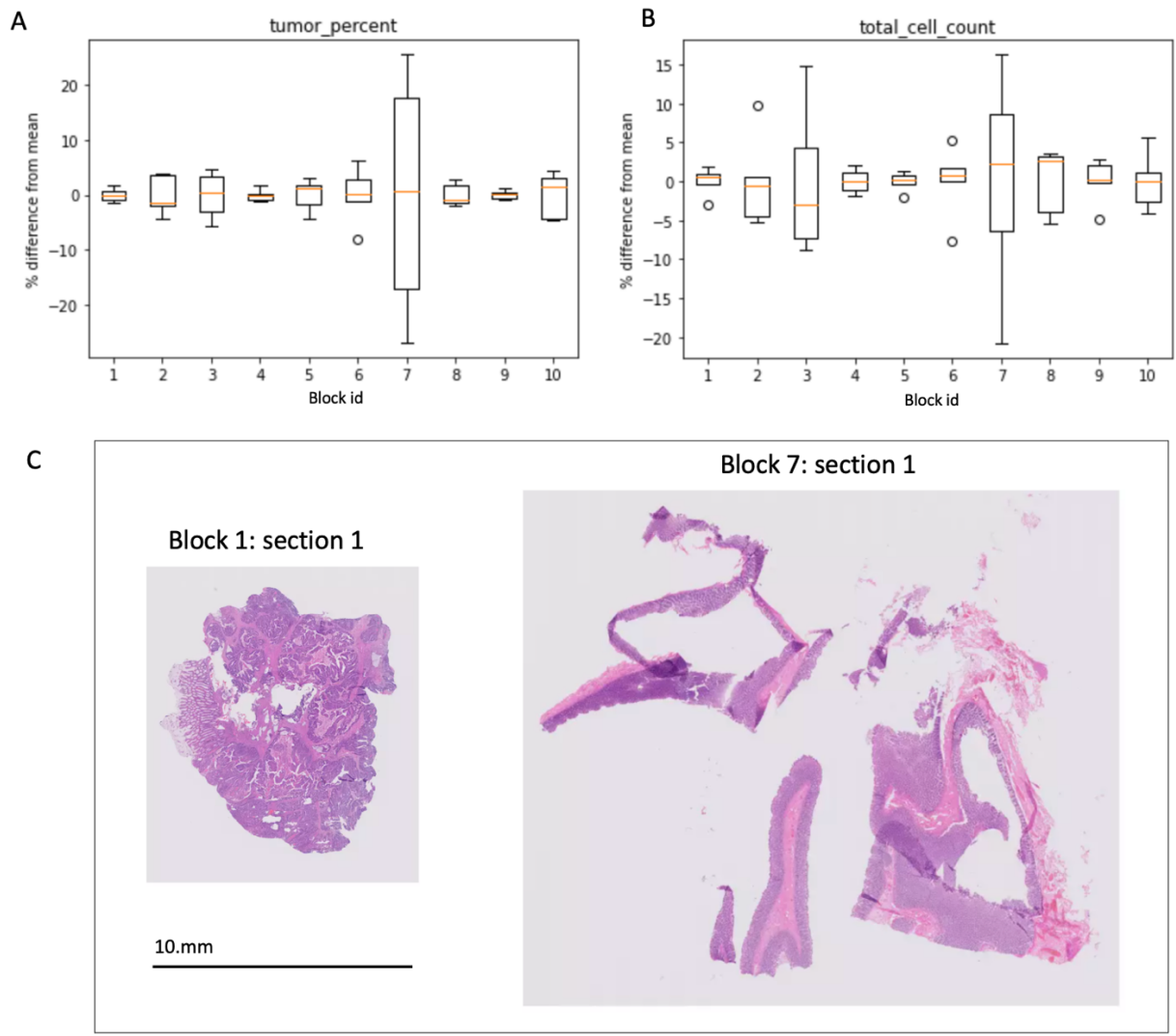

**Supplementary Figure 1. Study of variability across consecutive sections for tumor segmentation and cell segmentation algorithms.**
5 consecutive sections were taken from 10 different colorectal FFPE blocks. All blocks were primary colorectal tissue and resections. Each section is about 4-6 um thick, accounting for an approximately 25 um thick sampling area across the sections. Each section is HE stained and scanned (Philips UFS, Eindhoven The Netherlands). Tumor percent and total cell count was measured for each slide by running tumor segmentation and cell segmentation on each slide. Three slides were excluded from this analysis due to folds and artifacts. **A)** The difference between each slide's tumor % and the mean of all slides from the block (indicated by block id on x-axis) was scaled by 100/mean to obtain a percent difference for each slide. This is plotted for each of the 10 slides. Red lines represent median deviation, boxes represent 25th and 75th quartiles, and error bars represent standard deviation. **B)** Same as A) but for total cell count. For both tumor percent and cell count, we found that 8/10 of the blocks displayed variation in difference from the mean across sections within -5% to 5%. This confirms that for most slides the assumption of constant number of cells per slide is a fair approximation. **C)** Block 7, which has the largest variability in both tumor percent and total cell count between serial sections, consists of many fractured tissues. The remaining 10 slides look more similar to Block 1, which is a single resection tissue.

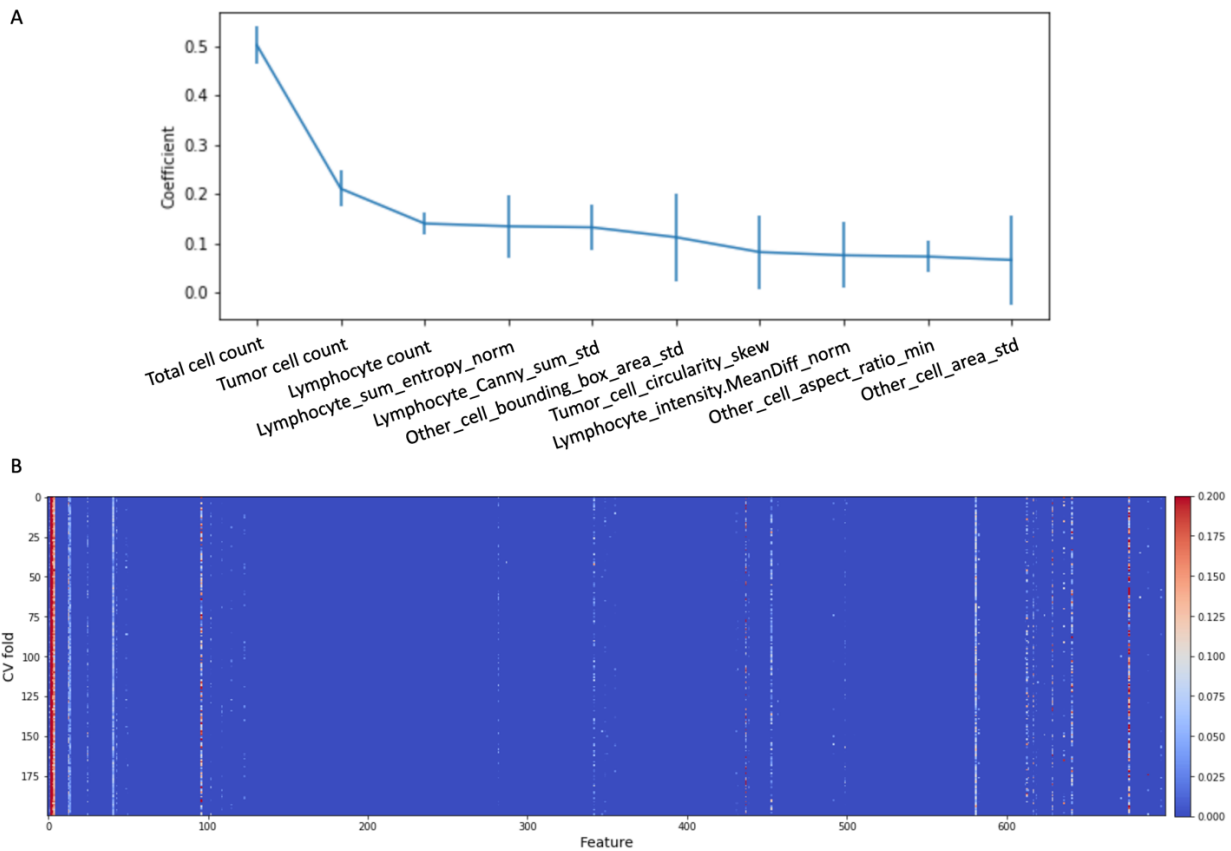

**Supplementary Figure 2. Feature importance and stability.**
**A)** Feature importance can be interpreted as the magnitude of linear model coefficient because features were normalized before model fitting. 200-fold cross validation using a 80/20 train/val split of the training set was performed using the best performing parameter combination (log-transform, L1 regularization strength 0.01). The linear model coefficients fit for each of the 200 folds were averaged and sorted in descending order. Shown are the mean and standard deviations of the top 10 coefficients. The total cell count is by far the most predictive feature, followed by the remaining cell count features and a mixture of tumor shape, cell shape, and cell texture features. **B)** Feature stability is demonstrated by plotting the magnitude of model coefficients for each feature per fold. The most predictive features have consistent coefficient magnitude across folds, as evidenced by vertical lines in the plot. For increased visibility, only the first 700 features are shown and the color scale for coefficient magnitude is cut off at 0.20.

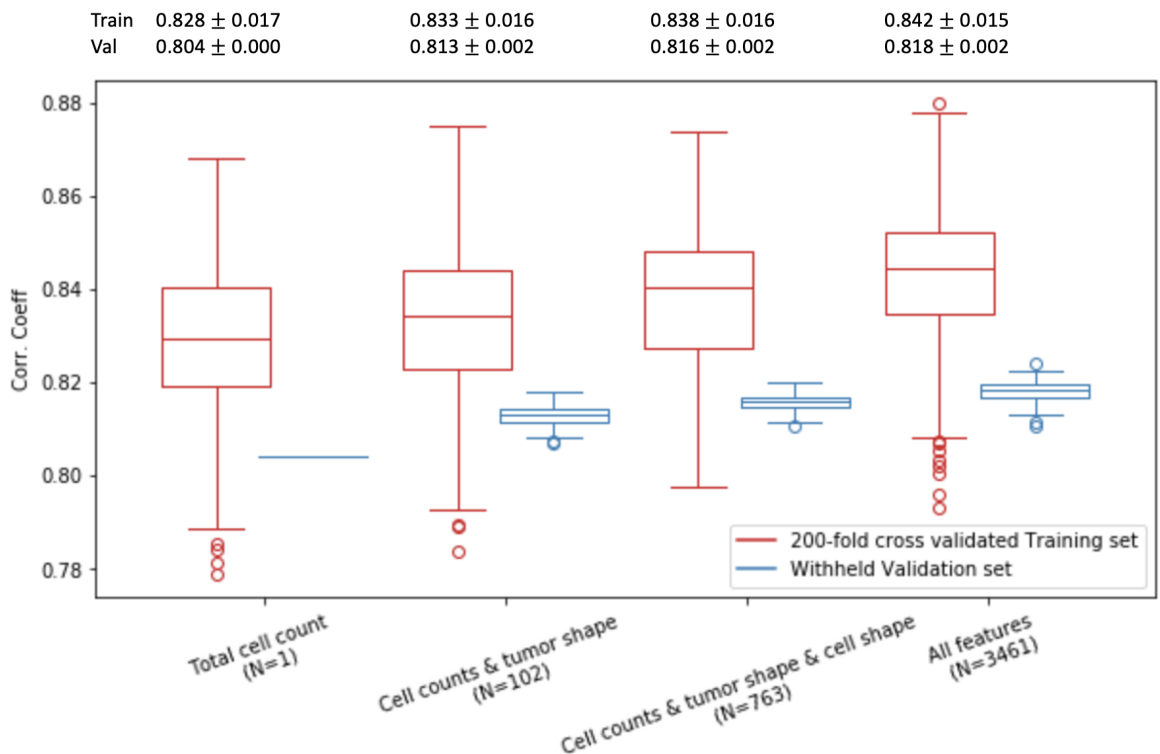

**Supplementary Figure 3. Correlation of predicted and true DNA yield increases with added imaging feature groups.**
Four groups with increasing numbers of features are considered. For each group 200-fold cross validation of the training set using a 80/20 train/val split is performed using optimal model parameters (red). For each fold the fit model is also evaluated on the withheld validation set (blue). The withheld validation set is the same for each fold, but the 20% of the training set used for cross validation varies randomly on each fold, hence the larger standard deviation. Train and val mean and standard deviation of correlation coefficient are shown above. Model parameters for each fold are fixed to the optimal combination obtained from parameter exploration study (log-transform, L1 regularization strength 0.01). Horizontal lines indicate median correlation coefficient, box width spans 25th to 75th percentile. Mean and standard deviation of correlation coefficient (*R*) for each feature group are shown above. For both training and withheld validation sets, the total cell count alone already correlates strongly with DNA yield per slide. This one-feature model has almost no variability on the validation set despite being fit to a different training set on each fold. Addition of feature groups steadily increases mean correlation coefficient for both training and validation sets while keeping standard deviation relatively constant.

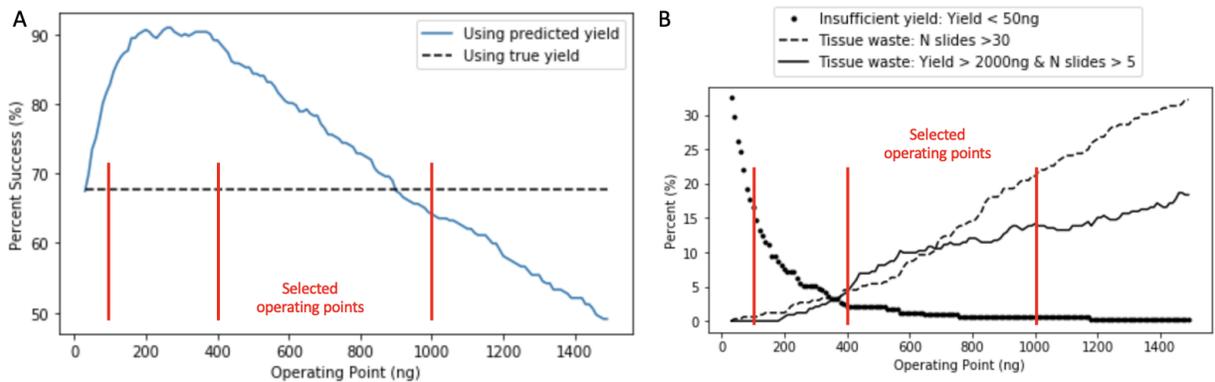

**Supplementary Figure 4. Operating point calibration on validation set**

We performed calibration of our model on the validation set (N = 332) to select three appropriate values of the target yield. For these simulations, Percent Success was defined as a yield within 60 – 2000 ng and < 30 slides scraped. Because validation set samples are archival, they already have extracted DNA yields, which we refer to as the "true yield". **A)** Percent Success using predicted yield on the validation set is simulated over a range of operating points (blue curve). The Percent Success for the true yield of the validation set is also shown (black dashed line). Collaborating pathologists chose three operating points (vertical red lines) to roughly span the range where Percent Success using the model is higher than the true Percent Success. **B)** Three failure modes are shown for the predicted yield vs operating point estimation in A. This demonstrates a trade-off between insufficient yield (where predicted mass is < 50ng) and tissue waste (where predicted number of slides is > 25 or yield is > 2000 ng and number of slides is > 5). Because costs incurred by insufficient quantity are higher than costs incurred by tissue waste, selection of the 1000 ng operating point was biased towards values that would minimize insufficient yield at the expense of tissue waste.

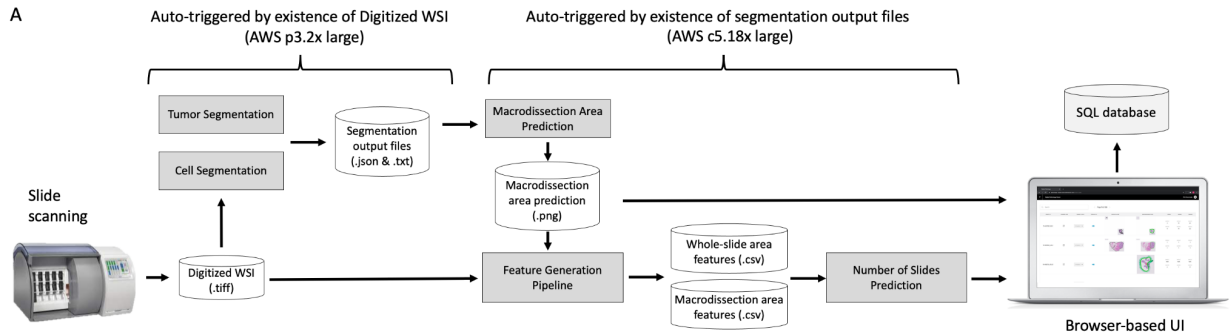

**Supplementary Figure 5. Model inference pipeline workflow for internal trial.**
**A)** During the internal trial a series of automated triggers are set up to run on AWS instances with AWS Lambda, executing all necessary model inference steps for AI-augmented path review. Scanning of a slide triggers tumor and cell segmentation inference, which are both run on an AWS p3.2xlarge instance. The segmentation outputs next trigger a series of computations which are run on an AWS m5.8xlarge instance: macrodissection area estimation, feature generation from both whole-slide and estimated macrodissection areas, and number of slides prediction. The Macrodissection area prediction and numbers of slides are output to a browser-based UI to be viewed by pathologists during review. After review the pathologist indicates the target yield in the UI. Model outputs and UI toggles are output to an SQL database, which is de-identified during data download. **B)** Example of model outputs displayed in browser based UI. Predicted macrodissection area is overlaid in green. Recommended numbers of slides to scrape are shown at the right for three target yields 100ng, 400ng, and 1000ng for both macrodissected and whole slide conditions.

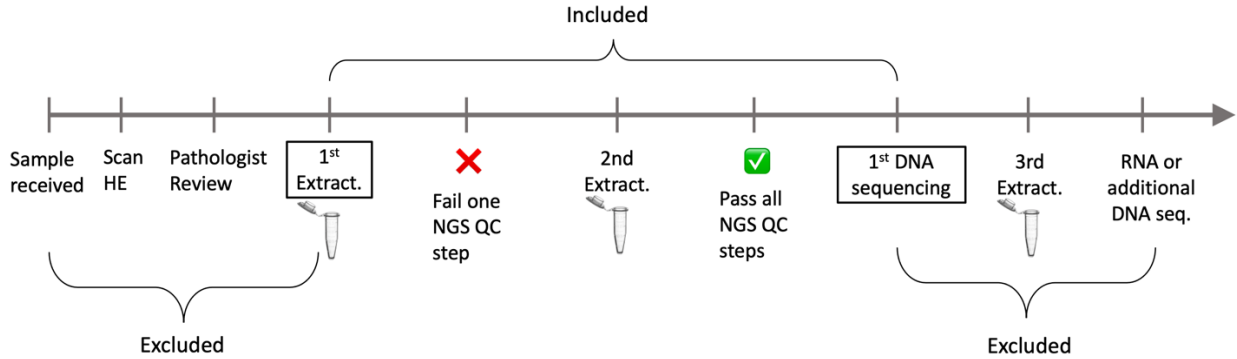

**Supplementary Figure 6. DNA sequencing time definition.**
Example timeline of a sample passing through our NGS sequencing workflow is shown. The DNA sequencing time (T-seq) is defined as the time delta between the first DNA sequencing event and the first DNA extraction attempt. This time range is chosen to minimize contributions of time periods over which we have no influence. Our NGS workflow includes QC at extraction, library preparation, hybridization, and sequencing steps. If any of these QC steps fail, there will be a re-extraction. In the schematic here we depict a 2nd extraction attempt following a QC failure, after which all QC steps are passed. A 3rd extraction, followed by RNA/DNA sequencing, occurs after the 1st DNA sequencing, but these are not included in the T-seq metric definition.

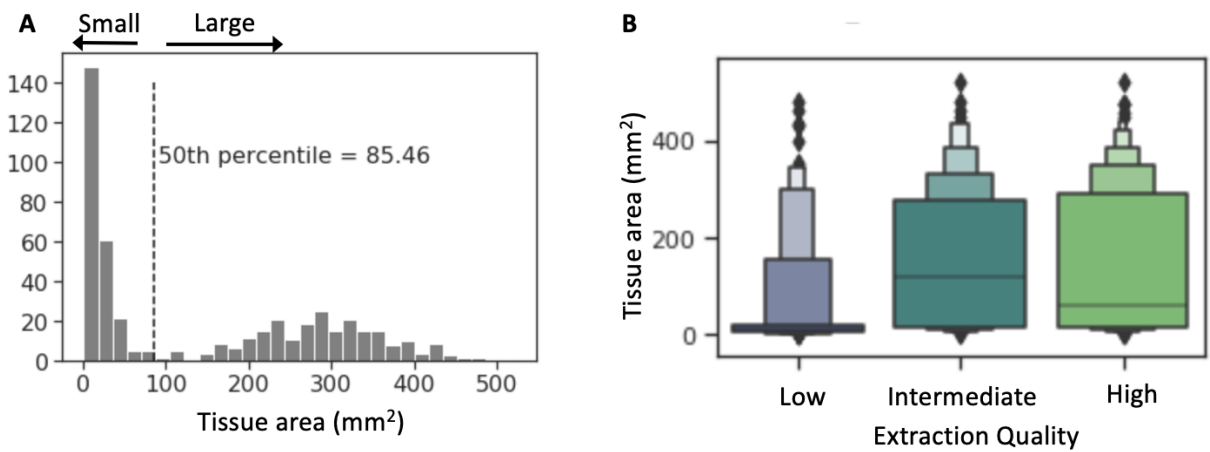

**Supplementary Figure 7. Details on Tissue area and extraction quality**
**A)** The distribution of tissue area for all samples enrolled into the trial is shown. The 50th percentile (85.46 mm$^2$) is used to split samples into large and small tissues. **B)** Boxen plot of tissue area vs extraction quality shows that low extraction quality samples tend to be smaller, while intermediate and high quality samples tend to be larger.

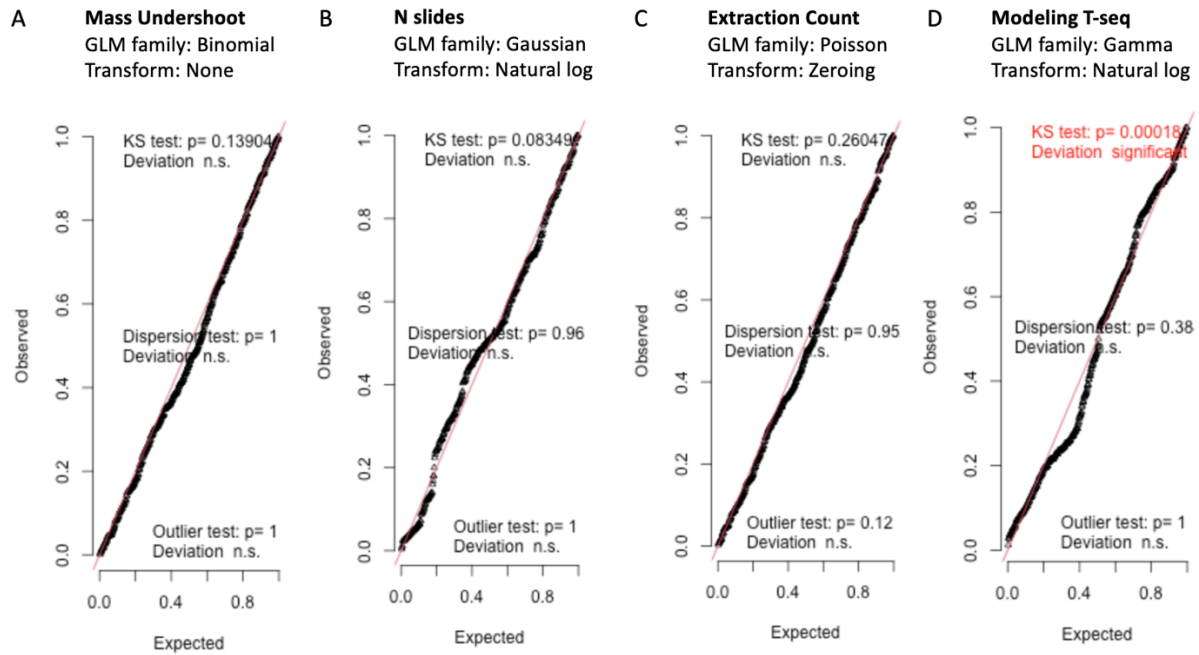

**Supplementary Figure 8. DHARMa residual diagnostics for GLM models of trial metrics**
GLMs were fit to four trial metrics (A-D) with three independent variables: trial cohort, extraction quality, and tissue area. The metric name, GLM family, and variable transformation used to model each metric are indicated above each plot. In order to characterize goodness-of-fit, simulated residuals were generated and plotted against expected residuals for each GLM family using the DHARMa package in R. For all modeled metrics, the expected and observed residuals closely follow the unity line, indicating well fit models. For each model, three goodness of fit tests were also performed, the Kolomogorov-Smirnov (KS) test, Dispersion test, and Outlier test. Each model passes all three tests, except for the Gamma family GLM modeling T-seq, which fails KS test.

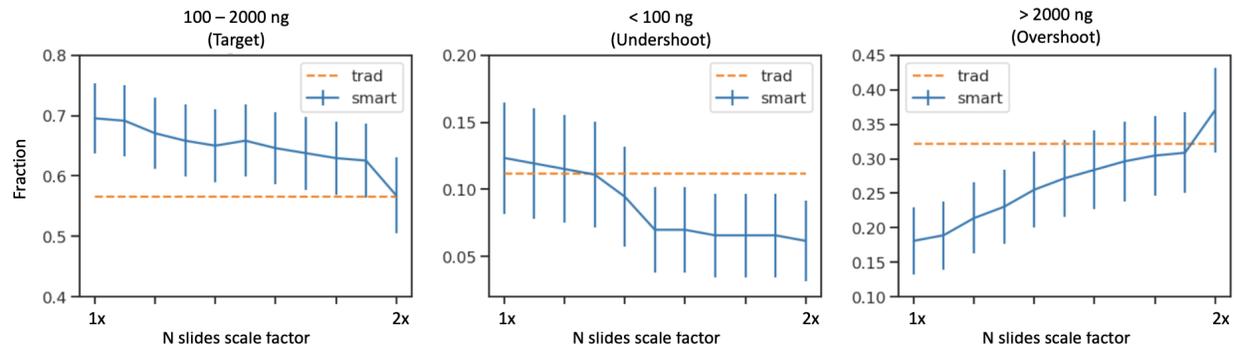

**Supplementary Figure 9. Simulated impact of scraping more slides.**

For each sample in the smart cohort the hypothetical DNA mass of the first extraction attempt was recomputed by scaling the measured DNA mass by the ratio of hypothetical N slides / actual N slides, where hypothetical N slides was increased from 1x to 2x times the number of the actual N slides. Because N slides is an integer, we round decimals down. For example, for 9 slides a 1.4x increase is 9*1.4 = 12.6 → 12 slides, but for 2 slides the 1.4x increase is still just 2 slides (2*1.4 = 2.8 → 2) The plots show the fraction of samples within the target (left), undershoot (middle), and overshoot (right) ranges plotted vs the hypothetical N slides scale factor. Errorbars represent 95% confidence intervals computed across the hypothetical DNA yields for all samples in the Smart cohort (N = 243). 1x on the x-axis corresponds directly to the true numbers presented in Results Figure 1A. Errorbars for trad cohort not shown. As the N slides scale factor is increased from 1x to 2x, the undershoot fraction reduces, while overshoot fraction increases. This suggests that we could improve the undershoot fraction at the expense of the overshoot fraction. This simulation makes two assumptions. First, it assumes that for each tissue, up to 2x the number of slides could have been scraped. For most samples this is true, but for some samples with limited tissue this assumption would fail. Second, it assumes linearity such that the DNA yield is the same for consecutive slides cut from the same FFPE block, which is a fair assumption on average for most samples.

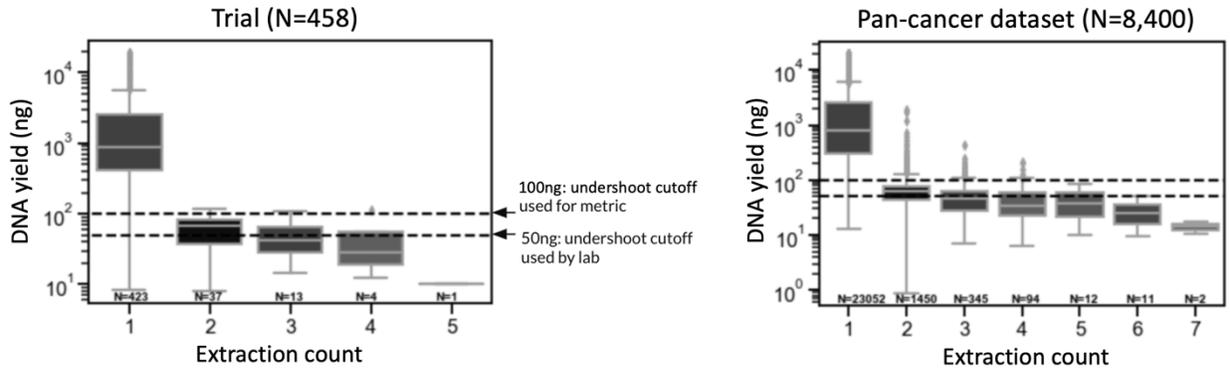

**Supplementary Figure 10. Relationship between DNA yield and N extractions**.
DNA yields under 100 ng almost always result in multiple (> 1) extraction attempts. (Left) Restricted to the trial data (colorectal only N = 476). (Right) Including a larger pan-cancer dataset (48, cancer types N = 8,400).

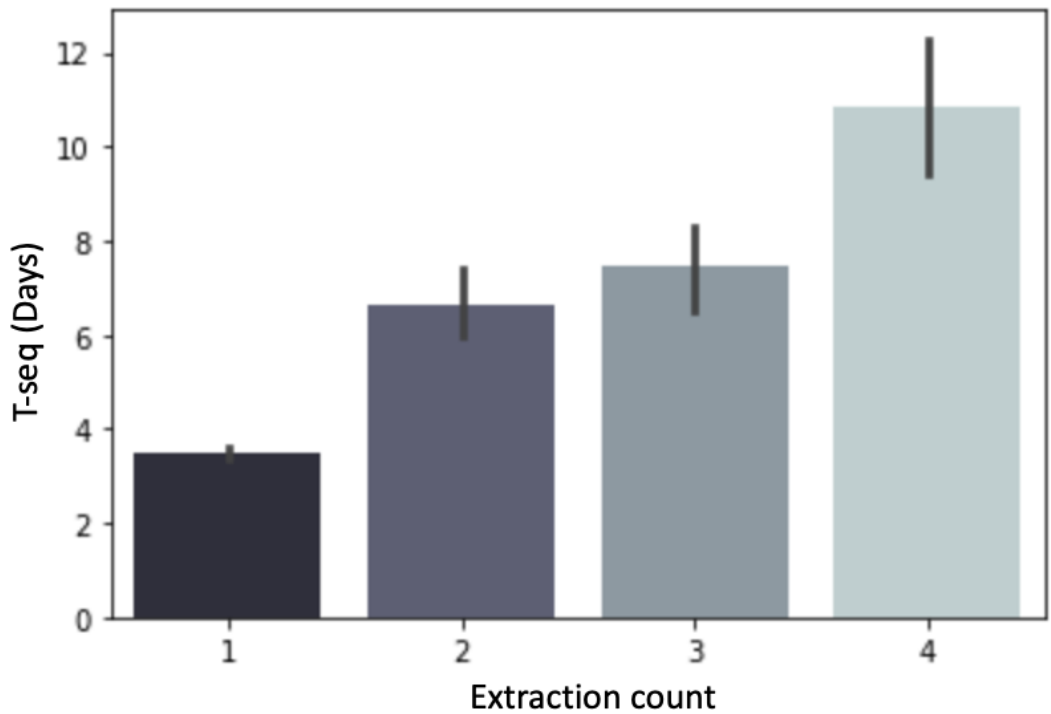

**Supplementary Figure 11. T-seq is strongly correlated with extraction count**

The extraction count for each sample in the internal trial data is plotted vs the T-seq for that sample. Bars show mean T-seq, error-bars show standard deviation.

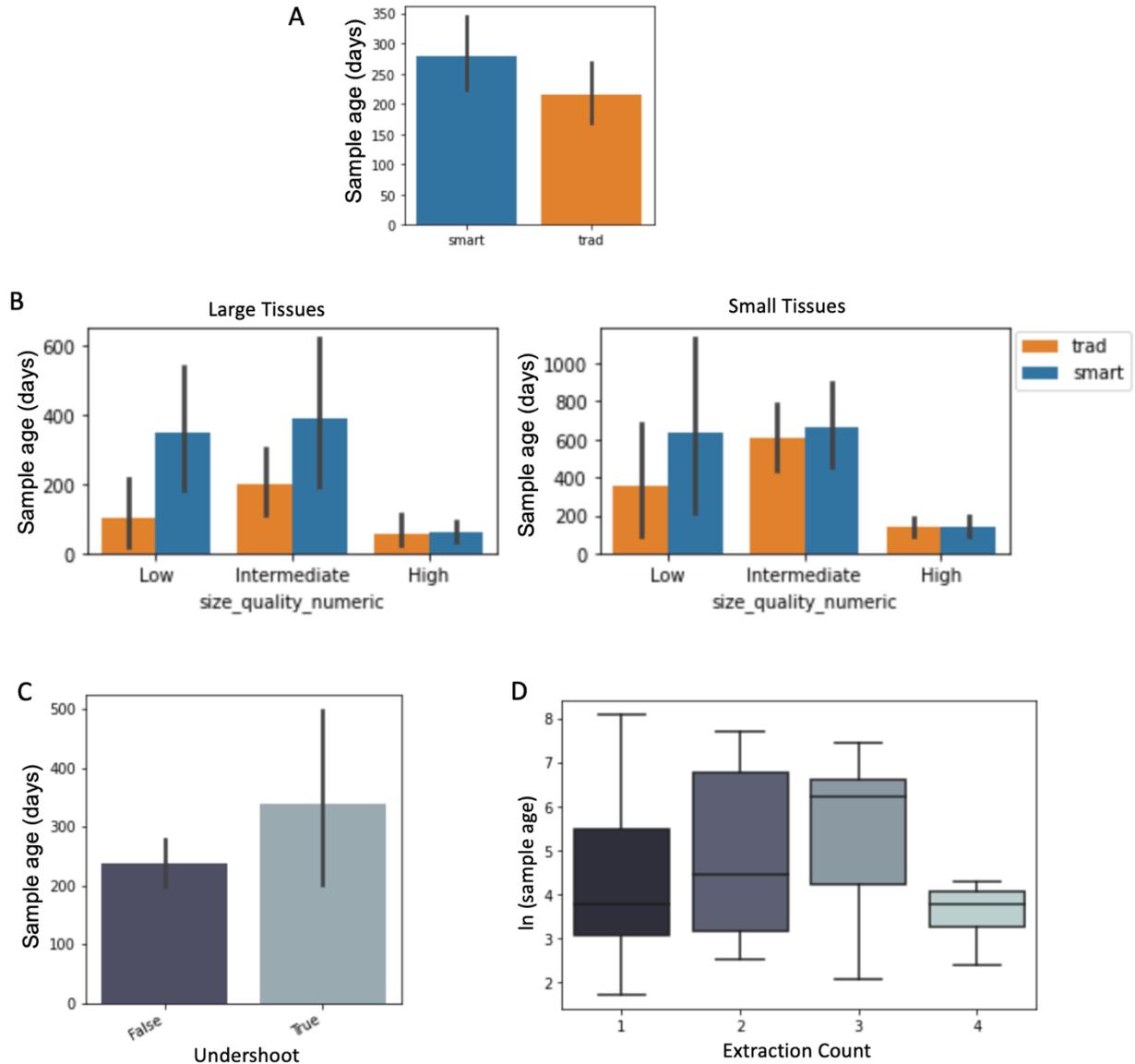

**Supplementary Figure 12. Sample age relationships with DNA mass undershoot, extraction count, and extraction quality.**
**A)** SmartPath cohort ("smart") samples are on average older than Trad cohort samples. **B)** When subsetted by large and small tissues, the oldest samples tend to be the smart cohort samples with intermediate quality. High quality samples are much younger than low and intermediate quality samples, regardless of subsetting. **C)** Undershoot samples (DNA mass < 100 ng) have higher mean sample age.

**D)** Older samples have more extraction attempts on average. Though the trend does not continue through N=4, only 4 samples in the entire dataset had 4 extraction attempts.

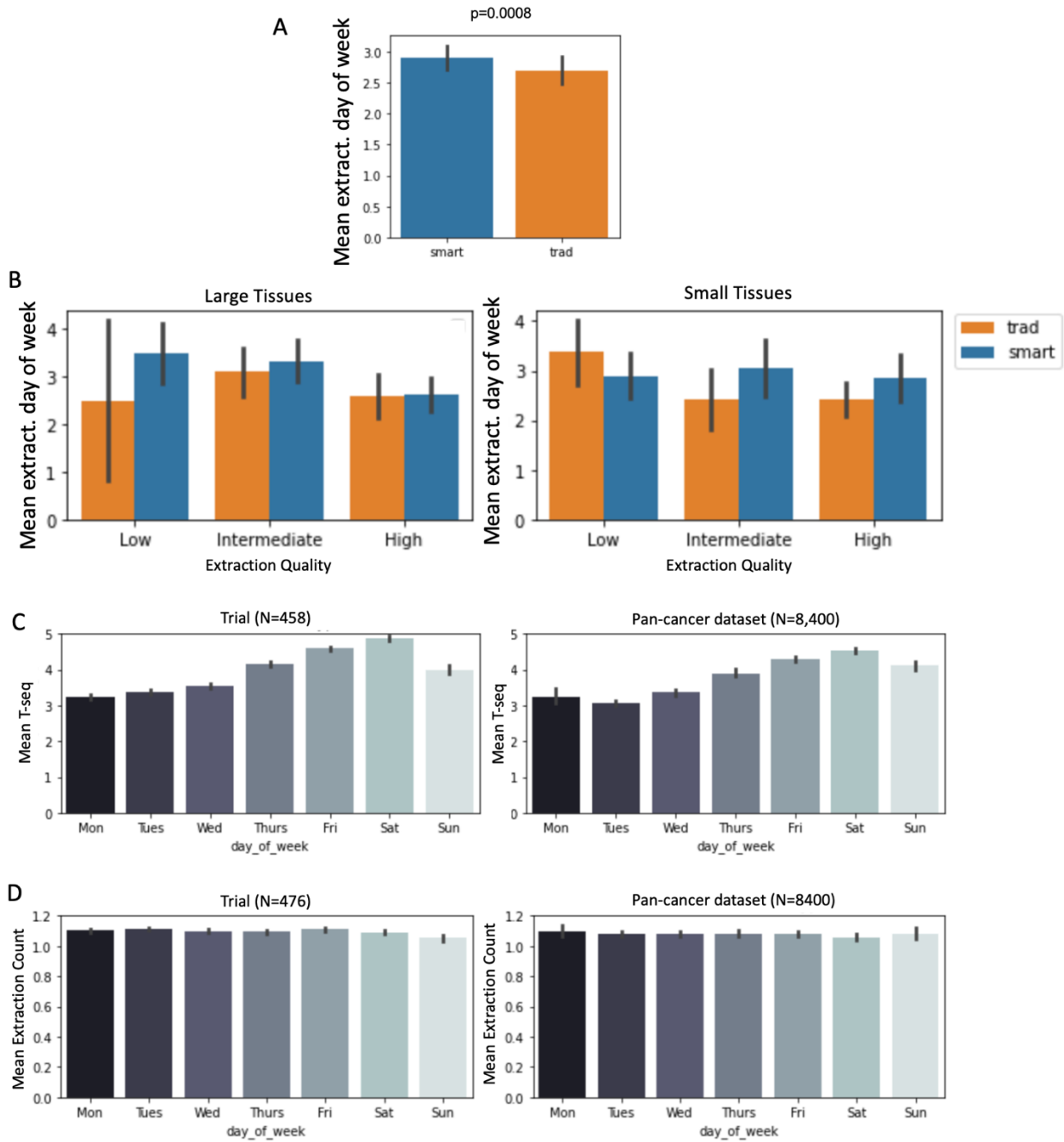

**Supplementary Figure 13. Extraction day of week impact on T-seq and extraction count metrics.**
**A)** Cohort imbalance for extraction day of week is shown. For this plot the day-of-week was represented numerically with Mon-Sun as 0-6. Chi-square p value computed from the contingency table is shown. **B)** Subsetting by tissue area and extraction quality. **C)** Mean T-seq is shown for samples whose first extraction attempt occurs on the indicated day of the week. The prominent increase of T-seq with day of week is due to the "weekend effect", where fewer personnel are available to process samples during the weekend. This effect is not due to a particular sampling of data, as it occurs equally in for samples enrolled into the internal trial (left) and a larger representation of our data, consisting of 8,400 samples from 48 different cancer types (right). **D)** Same, but for extraction count. No weekend effect is seen for extraction count.

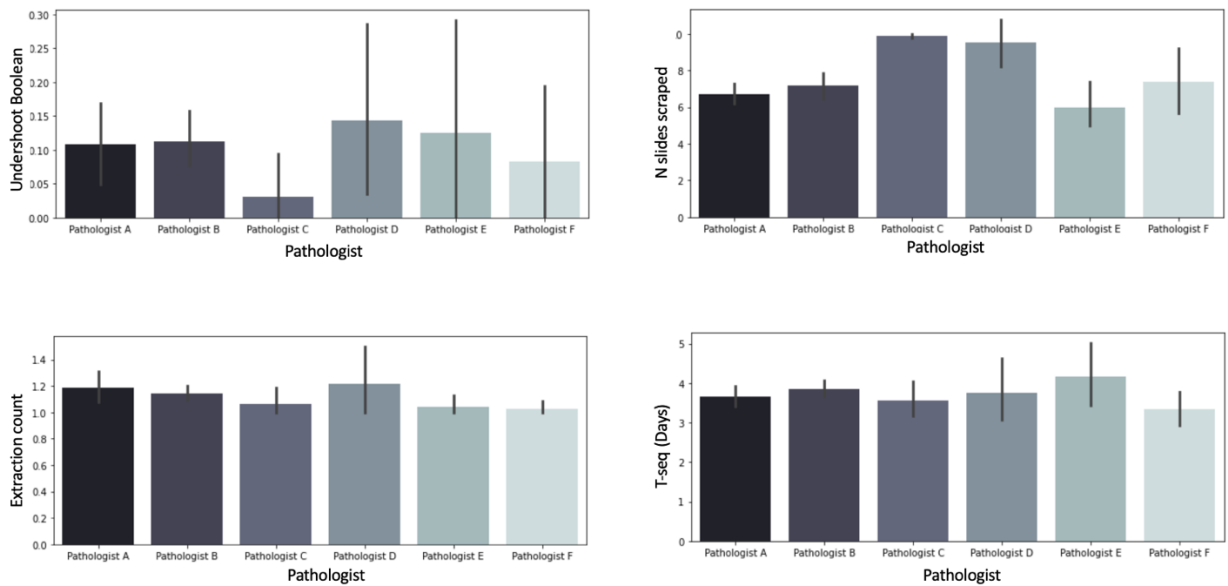

**Supplementary Figure 14. Pathologist effects on trial metrics**
Bar plots show the mean of the trial metrics when grouped by reviewing pathologists. Each metric is labeled on y-axis. Errorbars are 95% confidence intervals. Since undershoot boolean (upper left) is a binary value, the errorbars are large. Note that Pathologist A reviewed the largest fraction of Trad samples while Pathologist B reviewed the largest fraction of Smart samples.

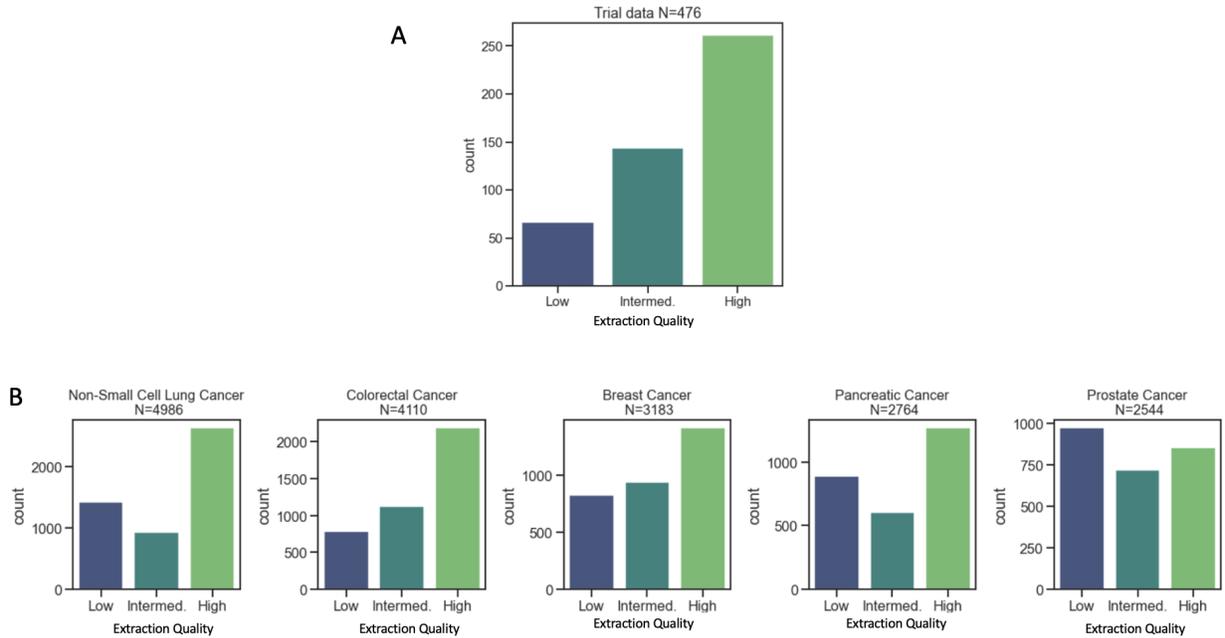

**Supplementary Figure 15. Extraction quality distributions for internal trial and other cancer types.
A)** Extraction quality distribution for internal trial dataset, which are all colorectal tumors. **B)** Extraction quality distributions for top 5 cancer types in our database. Cancer type and number of samples indicated in the title of each figure. The distributions of the trial data, only containing 478 samples (shown in A), is very similar to the larger distribution of colorectal data, containing 4,100 samples (shown in B), which indicates that the trial was a fair representation of the larger dataset. From the trial we found that AI-assistance is most likely to reduce extraction attempts for low quality samples, but the cancer type selected for the trial

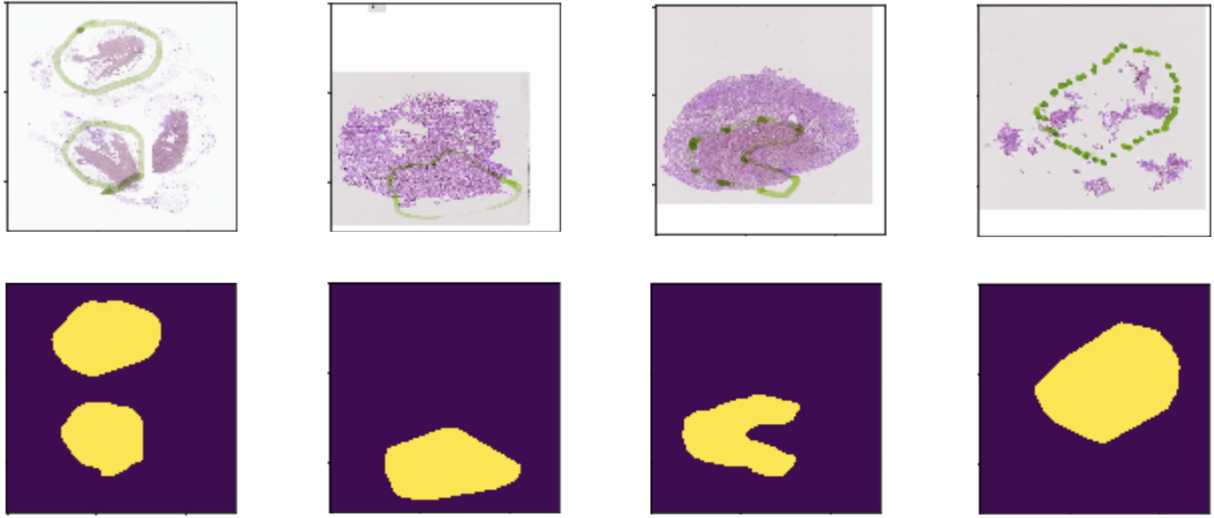

**Supplementary Figure 16. Marker area filling algorithm is robust to various marking patterns**
Ink on the slide is first detected by a trained U-Net model and then post-processed to produce a binary mask of the area enclosed by the marker. (left to right) Examples showing various marker patterns and the filled binary masks: multiple marked areas, marker contour doesn't fully close, marker contour contains concave area which must be excluded, dotted markings.

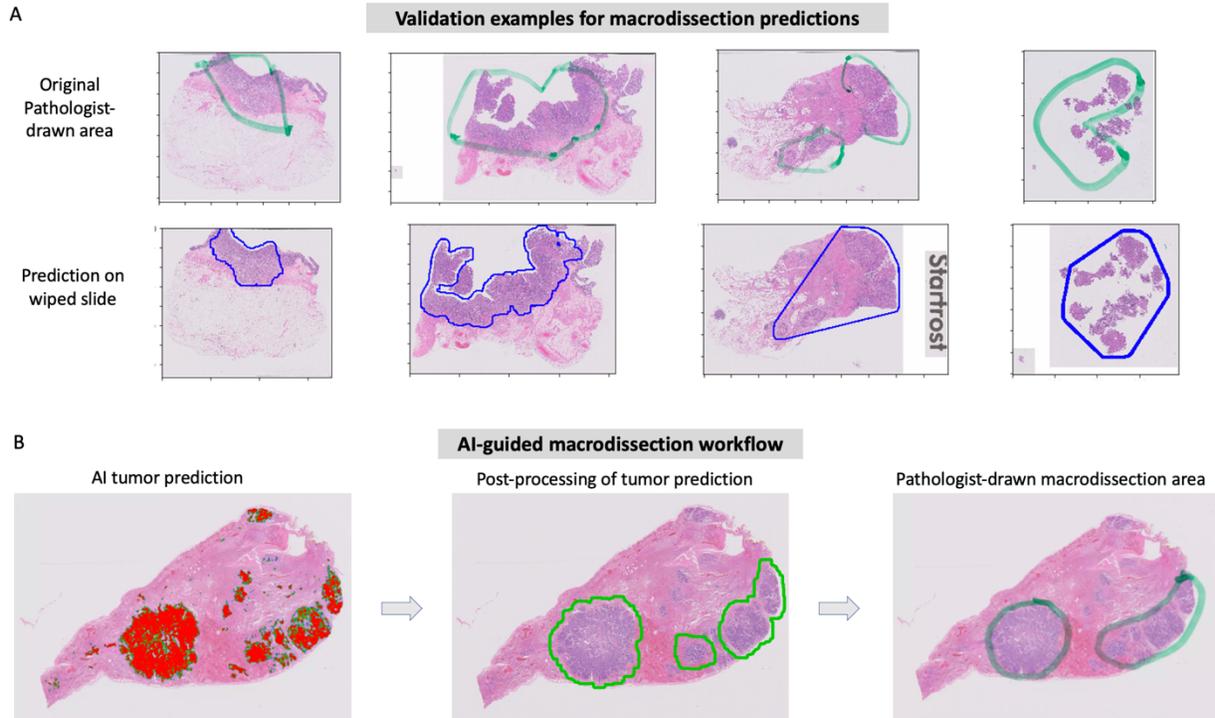

**Supplementary Figure 17. Macrodissection area estimation during inference: validation examples and workflow**
**A)** Side-by-side comparison of hand-drawn macrodissection areas (top row, green ink) and predicted macrodissection areas (bottom row, outlined in blue) made on the same slides after ink has been wiped. The two left-most cases represent slides with good agreement between hand-drawn and predicted areas, while the rightmost cases represent worse agreement. **B)** Illustration of AI-guided macrodissection estimation workflow during the trial. (Left) The colorectal tumor segmentation model predicts regions likely to be tumor. (Middle) The predicted area is post-processed to generate a macrodissection area contour. This image is what is presented to pathologists in a browser-based UI during pathology review. (Right) The pathologist draws an area onto the slide for the technician to later scrape for DNA extraction. The hand-drawn area will usually be smoother and larger than the predicted area, because the pathologist considers what is a reasonable shape for the technician to scrape.